\newif\ifJOURNAL
\JOURNALfalse
\newif\ifCONF
\CONFfalse
\newif\ifarXiv
\arXivfalse
\newif\ifWP
\WPfalse
\newif\ifFULL
\FULLfalse
\newif\ifLATIN
\LATINfalse

\arXivtrue

\LATINtrue		

\newif\ifnotCONF	
\notCONFtrue
\ifCONF\notCONFfalse\fi

\newif\ifnotarXiv	
\notarXivtrue
\ifarXiv\notarXivfalse\fi

\newif\ifTR		
\TRfalse
\ifarXiv\TRtrue\fi
\ifWP\TRtrue\fi
\newif\ifnotTR
\notTRtrue
\ifarXiv\notTRfalse\fi
\ifWP\notTRfalse\fi

\newif\ifnotLATIN	
\notLATINtrue
\ifLATIN\notLATINfalse\fi

\ifCONF
  \newcommand{\GTPVIII}{vovk/etal:2005AIStatslocal}
  \newcommand{\GTPX}{vovk/etal:2005ALT}
  \newcommand{\GTPXI}{GTP11arXiv-local}
  
  \newcommand{\GTPXIV}{vovk:2005ALT-GTP14}
  \newcommand{\GTPXVI}{GTP16arXiv-local}
  \newcommand{\AdamsFournier}{adams/fournier:2003}
\fi
\ifarXiv
  \newcommand{\GTPVIII}{GTP8arXiv}
  \newcommand{\GTPX}{GTP10arXiv}
  \newcommand{\GTPXI}{GTP11arXiv}
  
  \newcommand{\GTPXIV}{GTP14arXiv}
  \newcommand{\AdamsFournier}{adams/fournier:2003full}
\fi
\ifWP
  \newcommand{\GTPVIII}{GTP8}
  \newcommand{\GTPX}{GTP10}
  \newcommand{\GTPXI}{GTP11}
  
  \newcommand{\GTPXIV}{GTP14}
  \newcommand{\AdamsFournier}{adams/fournier:2003full}
\fi
\ifFULL
  \newcommand{\GTPVIII}{GTP8arXiv}
  \newcommand{\GTPX}{GTP10arXiv}
  \newcommand{\GTPXI}{GTP11arXiv}
  
  \newcommand{\GTPXIV}{GTP14arXiv}
  \newcommand{\AdamsFournier}{adams/fournier:2003full}
\fi

\ifnotLATIN
  \newcommand{\Gurary}{gurary:1967}
  \newcommand{\Liokumovich}{liokumovich:1973}
  \newcommand{\LiptserShiryaev}{liptser/shiryaev:1974}
  \newcommand{\Nikolsky}{nikolsky:1961}
\fi
\ifLATIN
  \newcommand{\Gurary}{gurary:1967latin}
  \newcommand{\Liokumovich}{liokumovich:1973latin}
  \newcommand{\LiptserShiryaev}{liptser/shiryaev:1974latin}
  \newcommand{\Nikolsky}{nikolsky:1961latin}
\fi

\ifCONF
\documentclass{llncs}
\usepackage{amsmath,amsfonts,amssymb,latexsym,graphicx}
\newcommand{\Extra}[1]{}
\fi

\ifarXiv
\documentclass{article}
\usepackage{amsmath,amsfonts,amssymb,latexsym,graphicx}
\newcommand{\Extra}[1]{}
\fi

\ifWP
\documentclass[toc]{gtarticle}
\usepackage{amsmath,amsfonts,amssymb,latexsym,epsfig,graphicx}
\renewcommand{\Extra}[1]{#1}
\fi

\ifFULL
\documentclass{article}
\usepackage{amsmath,amsfonts,amssymb,latexsym,color,epigraph,graphicx}
\newcommand{\Extra}[1]{\red{#1}}
\fi

\allowdisplaybreaks[3]

\newcommand{\red}[1]{\textcolor{red}{#1}}

\newcommand{\bluebegin}{\begingroup\color{blue}}
\newcommand{\blueend}{\endgroup}
\newcommand{\redbegin}{\begingroup\color{red}}
\newcommand{\redend}{\endgroup}

\newcommand{\Vladimir}{Vladimir}
\newcommand{\DOT}{.}

\newcommand{\st}{\mathrel{\!|\!}}
\newcommand{\D}{\,\mathrm{d}}
\newcommand{\dd}{\mathrm{d}}

\newcommand{\diam}{\mathop{\rm diam}\nolimits}
\newcommand{\sign}{\mathop{\rm sign}\nolimits}

\newcommand{\risk}{\mathop{\mathrm{risk}}\nolimits}

\newcommand{\K}{\mathcal{K}}		
\newcommand{\kkk}{\mathbf{k}}		
\newcommand{\ccc}{\mathbf{c}}		

\newcommand{\FFF}{\mathcal{F}}		

\ifnotCONF
  \newcommand{\bbbp}{\mathbb{P}}		
\fi
\ifCONF
  \renewcommand{\bbbp}{\mathbb{P}}		
\fi
\newcommand{\Prob}{\mathop{\bbbp}\nolimits}

\ifnotCONF
\newcommand{\bbbr}{\mathbb{R}}		
\newtheorem{lemma}{Lemma}
\newtheorem{proposition}{Proposition}
\newtheorem{corollary}{Corollary}
\newtheorem{theorem}{Theorem}
\newenvironment{proof}
  {\trivlist\item[\hskip\labelsep\textbf{Proof}]}
  {\endtrivlist}
\fi

\newenvironment{Proof}[1]
  {\trivlist\item[\hskip\labelsep\textit{Proof #1:}]}
  {\endtrivlist}
\newcommand{\boxforqed}{\rule{.3em}{1.5ex}}
\newcommand{\qedtext}{\unskip\nobreak\hfil
  \penalty50\hskip1em\null\nobreak\hfil\boxforqed
  \parfillskip=0pt\finalhyphendemerits=0\endgraf}
\newcommand{\qedmath}{\tag*{\boxforqed}}
\newenvironment{remark*}
  {\trivlist\item[\hskip\labelsep{\bfseries Remark}]\relax}
  {\endtrivlist}

\newlength{\IndentI}
\newlength{\IndentII}
\newlength{\IndentIII}
\newlength{\WidthI}
\newlength{\WidthII}
\newlength{\WidthIII}
\ifCONF
\title{Competing with wild prediction rules}
\author{Vladimir Vovk}
\institute{Computer Learning Research Centre,
  Department of Computer Science\\
  Royal Holloway, University of London,
  Egham, Surrey TW20 0EX, UK\\
  \email{vovk@cs.rhul.ac.uk}}
\fi

\ifarXiv
\title{Competing with wild prediction rules}
\author{Vladimir Vovk\\
\texttt{vovk{\rm@}cs.rhul.ac.uk}\\
\texttt{http://vovk.net}}
\fi

\ifWP
\title{Competing with wild prediction rules}
\author{Vladimir Vovk}

\fi

\ifFULL
\title{Competing with wild prediction rules}
\author{Vladimir Vovk\\
\texttt{vovk{\rm@}cs.rhul.ac.uk}\\
\texttt{http://vovk.net}}
\fi

\begin{document}
\maketitle
\begin{abstract}
  We consider the problem of on-line prediction
  competitive with a benchmark class of continuous but highly irregular prediction rules.
  It is known that if the benchmark class is a reproducing kernel Hilbert space,
  there exists a prediction algorithm
  whose average loss over the first $N$ examples
  does not exceed the average loss of any prediction rule in the class
  plus a ``regret term'' of $O(N^{-1/2})$.
  The elements of some natural benchmark classes, however,
  are so irregular
  that these classes are not Hilbert spaces.
  In this paper we develop Banach-space methods
  to construct a prediction algorithm
  with a regret term of $O(N^{-1/p})$,
  where $p\in[2,\infty)$ and $p-2$ reflects the degree to which the benchmark class
  fails to be a Hilbert space.
\end{abstract}

\section{Introduction}
\label{sec:introduction}

For simplicity, in this introductory section we only discuss
the problem of predicting labels $y_n$ of objects $x_n\in[0,1]$
(this will remain our main example throughout the paper).
In this paper we are mainly interested in extending the class
of the prediction rules our algorithms are competitive with;
in other respects,
our assumptions are rather restrictive.
For example, we always assume that the labels $y_n$
are bounded in absolute value by a known positive constant $Y$
and only consider the problem of square-loss regression\ifnotCONF\
(some ideas for extension to a wider range of loss functions
can be found in \cite{\GTPXIV})\fi.

Standard methods allow one to construct
a ``universally consistent'' on-line prediction algorithm,
i.e.,
an on-line prediction algorithm whose average loss over the first $N$ examples
does not exceed the average loss of any continuous prediction rule
plus $o(1)$.
(Such methods were developed in, e.g., \cite{cesabianchi/long/warmuth:1996},
\cite{kivinen/warmuth:1997},
and, especially, \cite{auer/etal:2002}, \S3.2;
for an explicit statement see \cite{\GTPXI}.)
More specifically, for any reproducing kernel Hilbert space (RKHS) on $[0,1]$
one can construct an on-line prediction algorithm
whose average loss does not exceed that of any prediction rule in the RKHS
plus $O(N^{-1/2})$;
choosing a universal RKHS (\cite{steinwart:2001}, Definition 4)
gives universal consistency.
In this paper we are interested in extending the latter result,
which is much more specific than the $o(1)$ provided by universal consistency,
to wider benchmark classes of prediction rules.
First we discuss limitations of RKHS as benchmark classes.

\begin{figure}[b]
  \hfil%
  \begin{minipage}[t]{.3\linewidth}
    \begin{center}$h=0.2$\end{center}
    \begin{center}
      \makebox{\includegraphics[width=\linewidth]{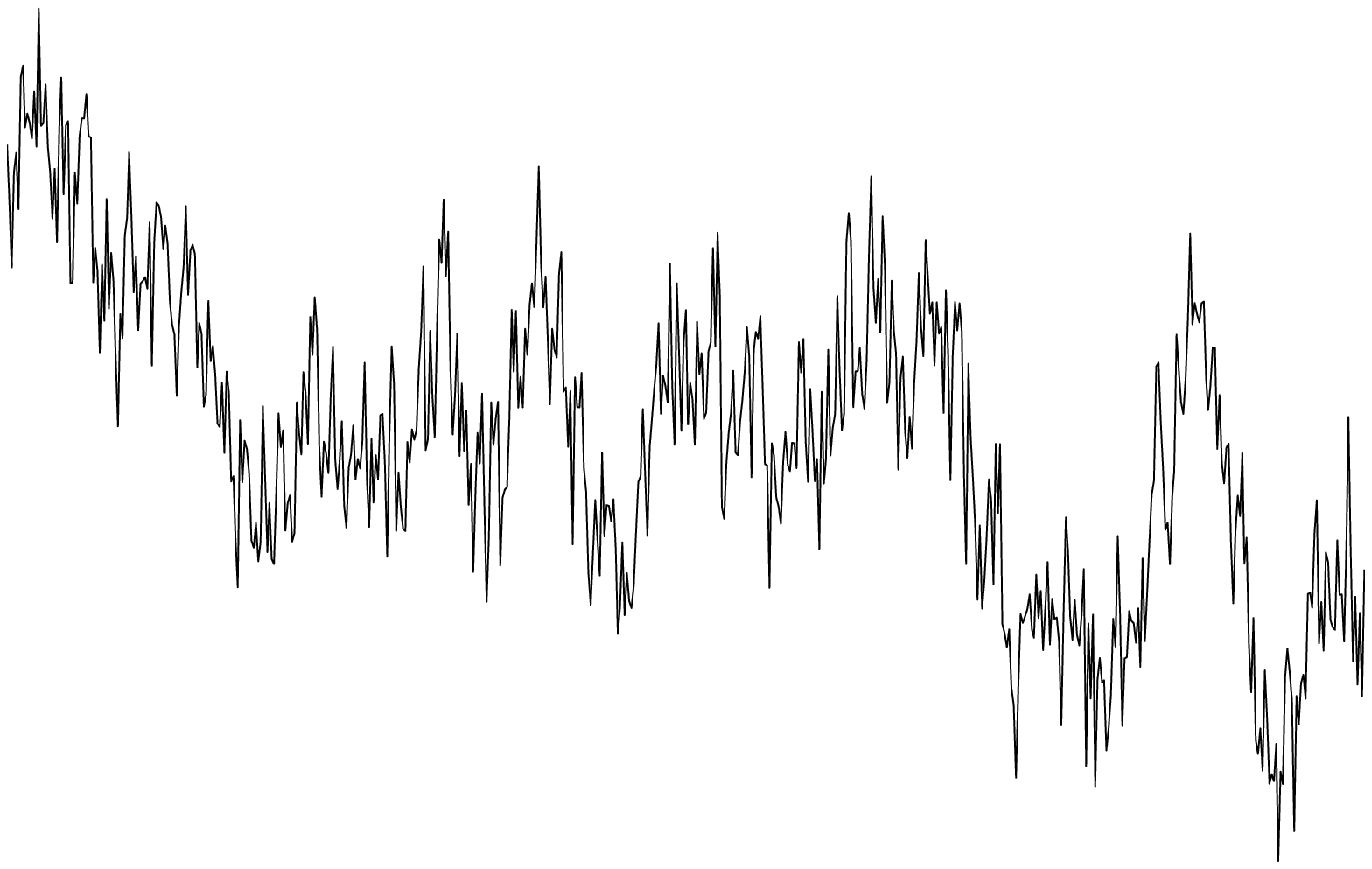}}
    \end{center}
  \end{minipage}\hfil
  \begin{minipage}[t]{.3\linewidth}
    \begin{center}$h=0.5$\end{center}
    \begin{center}
      \makebox{\includegraphics[width=\linewidth]{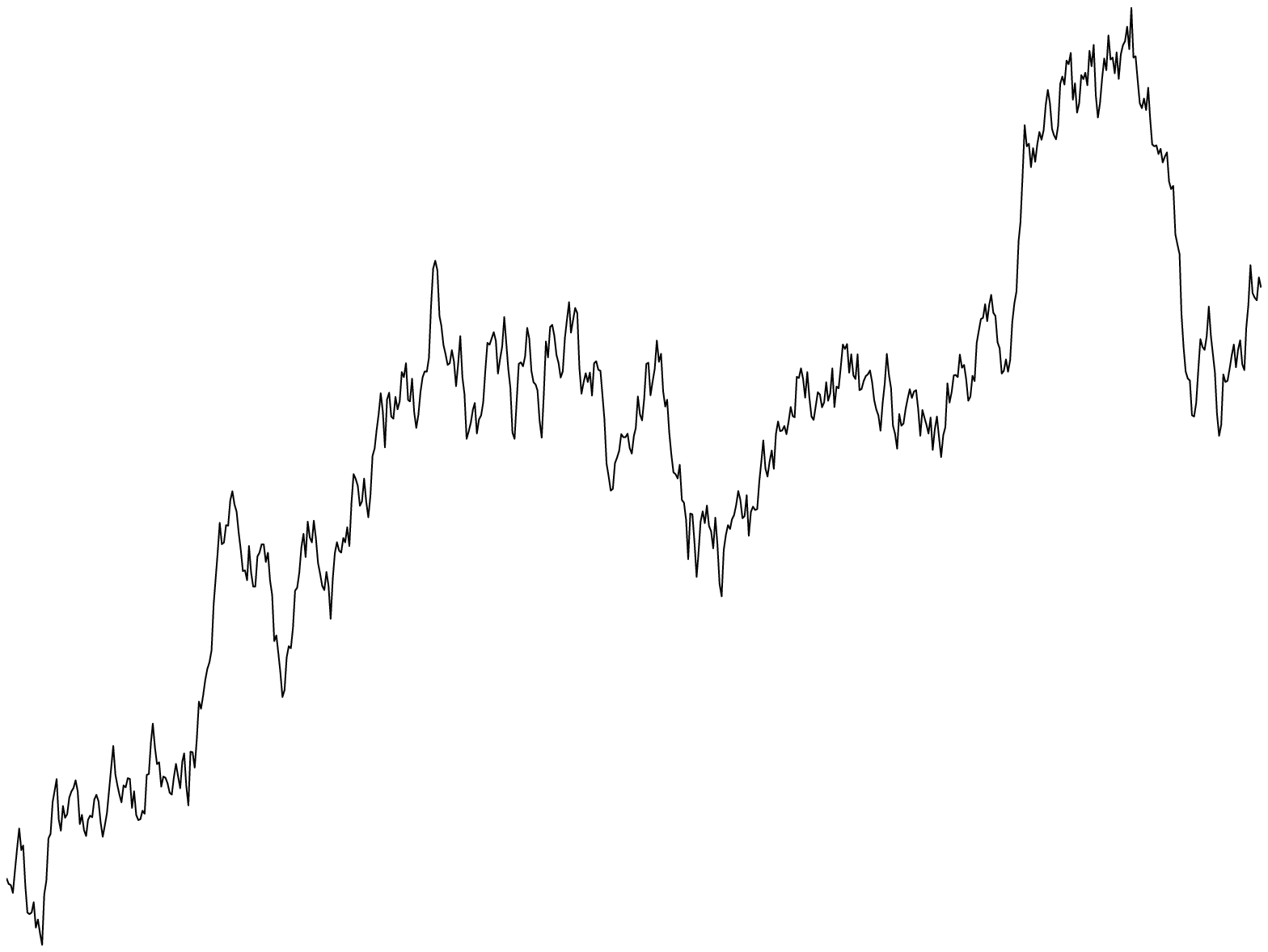}}
    \end{center}
  \end{minipage}\hfil
  \begin{minipage}[t]{.3\linewidth}
    \begin{center}$h=0.8$\end{center}
    \begin{center}
      \makebox{\includegraphics[width=\linewidth]{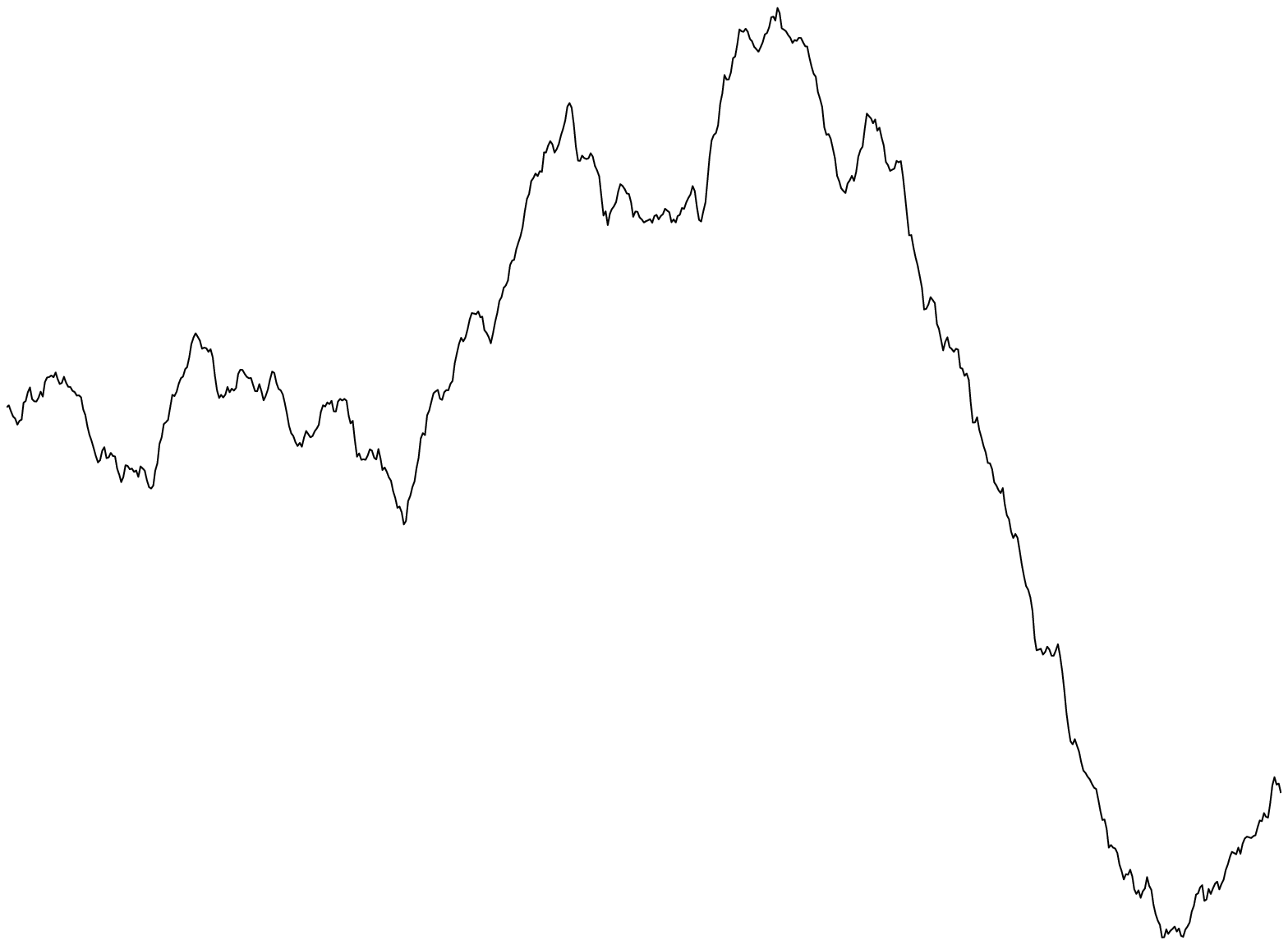}}
    \end{center}
  \end{minipage}
  \caption{Functions with H\"older exponent $h$
    for three different values of $h$.}
  \label{fig:fbm}
\end{figure}

The regularity of a prediction rule $D$
can be measured by its ``H\"older exponent'' $h$,
which is informally defined by the condition
that $\left| D(x+dx)-D(x) \right|$ scale
as $\left| dx \right|^h$ for small $\left| d x\right|$.
The most regular continuous functions are those of classical analysis:
say, piecewise differentiable with bounded derivatives.
For such functions the H\"older exponent is $1$.
\ifnotCONF
  Familiar examples are $x\mapsto\sin x$ and $x\mapsto\left|x-1/2\right|$.
\fi
Functions much less regular than those of classical analysis
are ubiquitous in probability theory:
for example, typical trajectories of the Brownian motion
(more generally,
of non-degenerate diffusion processes) 
have H\"older exponent $1/2$.
Functions with other H\"older exponents $h\in(0,1)$
can be obtained as typical trajectories of the fractional Brownian motion.
Three examples with different values of $h$ are shown in Figure \ref{fig:fbm}.

The intuition behind the informal notion of a function with H\"older exponent $h$
will be captured using function spaces known as Sobolev spaces.
Roughly, the Sobolev spaces $W^{s,p}([0,1])$
(defined formally in the next section),
where $p\in(1,\infty]$, $s\in(0,1)$, and $s>1/p$,
can be regarded as different ways of formalizing the notion of a function on $[0,1]$
with H\"older exponent $h>s$.

The most familiar Sobolev spaces are the H\"older spaces $W^{s,\infty}([0,1])$,
consisting of the functions $f$ satisfying
$
  \lvert f(x)-f(y) \rvert
  =
  O
  \left(
    \lvert x-y \rvert^s
  \right)
$.
The H\"older spaces are nested,
$W^{s,\infty}([0,1])\subset W^{s',\infty}([0,1])$
when $s'<s$.
\ifnotCONF (That all H\"older spaces are very different can be seen
from the fact that typical trajectories of the fractional Brownian motion
$B^{(h)}$, defined in \S\ref{sec:stochastic},
are in $W^{s,\infty}([0,1])$ for $s<h$
and outside $W^{s,\infty}([0,1])$ for $s>h$.) \fi
As we will see in a moment,
the standard Hilbert-space methods only work
for $W^{s,\infty}([0,1])$ with $s>1/2$ as benchmark classes;
our goal is to develop methods that would work for smaller $s$ as well.

The spaces $W^{s,\infty}([0,1])$ are rather awkward analytically
and even poorly reflect the intuitive notion of H\"older exponent:
they are defined in terms of 
$
  \sup_{x,y}
  \lvert f(x)-f(y) \rvert / \lvert x-y \rvert^s
$,
and so $f$'s behavior in the neighborhood of a single point can disqualify it
from being a member of $W^{s,\infty}([0,1])$.
Replacing $\sup$ with the mean (in the sense of $L^p$)
w.r.\ to a natural ``almost finite'' measure gives
the Sobolev spaces $W^{s,p}([0,1])$ for $p<\infty$.
Results for the case $p<\infty$ immediately carry over to $p=\infty$
since, as we will see in the next section,
$W^{s,\infty}([0,1])\subseteq W^{s',p}([0,1])$
whenever $s'<s$;
$s'$ can be arbitrarily close to $s$.

All Sobolev spaces (including the H\"older spaces)
are Banach spaces,
but $W^{s,2}([0,1])$ are also Hilbert spaces
and, for $s>1/2$, even RKHS.
Therefore,
they are amenable to the standard methods
(see the papers mentioned above;
the exposition of \cite{\GTPXI} is especially close to that of this paper,
although we wrote $H^s$ instead of $W^{s,2}$ in \cite{\GTPXI}).

The condition $s>1/p$ appears indispensable in the development of the theory
(cf.\ the reference to the Sobolev imbedding theorem
in the next section).
Since this paper concentrates on the irregular end of the Sobolev spectrum,
$s<1/2$,
instead of the Hilbert spaces $W^{s,2}([0,1])$
we now have to deal with the Banach spaces
$W^{s,p}([0,1])$ with $p\in(2,\infty)$,
which are not Hilbert spaces.
The necessary tools are developed in \S\S\ref{sec:geometry}--\ref{sec:proof}.

The methods of \cite{\GTPXI} relied on the perfect shape
of the unit ball in a Hilbert space.
If $p$ is not very far from $2$,
the unit ball in $W^{s,p}$ is not longer perfectly round
but still convex enough to allow us to obtain similar results by similar methods.
In principle,
the condition $s>1/p$ is not longer an obstacle to coping with any $s>0$:
by taking a large enough $p$ we can reach arbitrarily small $s$.
However, the quality of prediction (at least as judged by our bound)
will deteriorate:
as we will see (Theorem \ref{thm:main} in the next section),
the average loss of our prediction algorithm
does not exceed that of any prediction rule in $W^{s,p}([0,1])$
plus $O(N^{-1/p})$.
(This gives a regret term of $O(N^{-s+\epsilon})$
for the prediction rules in $W^{s,\infty}([0,1])$,
where $s<1/2$ and $\epsilon>0$.)


\section{Main result}
\label{sec:main}

We consider the following perfect-information prediction protocol:

\bigskip

\parshape=5
\IndentI  \WidthI
\IndentII \WidthII
\IndentII \WidthII
\IndentII \WidthII
\IndentI  \WidthI
\noindent
FOR $n=1,2,\dots$:\\
  Reality announces $x_n\in\mathbf{X}$.\\
  Predictor announces $\mu_n\in\bbbr$.\\
  Reality announces $y_n\in[-Y,Y]$.\\
END FOR.

\bigskip

\noindent
At the beginning of each round $n$ Predictor is given an object $x_n$
whose label is to be predicted.
The set of \emph{a priori} possible objects,
the \emph{object space},
is denoted $\mathbf{X}$;
we always assume $\mathbf{X}\ne\emptyset$.
After Predictor announces his prediction $\mu_n$ for the object's label
he is shown the actual label $y_n\in[-Y,Y]$.
We consider the problem of regression, $y_n\in\bbbr$,
assuming an upper bound $Y$ on $\left|y_n\right|$.
The pairs $(x_n,y_n)$ are called \emph{examples}.

Predictor's loss on round $n$ is measured by
$\left(y_n-\mu_n\right)^2$,
and so his average loss after $N$ rounds of the game is
$
  \frac1N
  \sum_{n=1}^N
  \left(
    y_n-\mu_n
  \right)^2
$.
His goal is to have
\begin{equation*}
  \frac1N
  \sum_{n=1}^N
  \left(
    y_n-\mu_n
  \right)^2
  \lessapprox
  \frac1N
  \sum_{n=1}^N
  \left(
    y_n-D(x_n)
  \right)^2
\end{equation*}
($\lessapprox$ meaning ``is less than or approximately equal to'')
for each prediction rule $D:\mathbf{X}\to\bbbr$
that is not ``too wild''.

\subsection*{Main theorem}

Our main theorem will be fairly general
and applicable to a wide range of Banach function spaces.
Its implications for Sobolev spaces will be explained after its statement.

Let $U$ be a Banach space and
$
  S_{U}
  :=
  \left\{
    u\in U
    \st
    \left\|
      u
    \right\|_{U}
    =
    1
  \right\}
$
be the unit sphere in $U$.
Our methods are applicable only to Banach spaces
whose unit spheres do not have very flat areas;
a convenient measure of rotundity of $S_U$
is Clarkson's \cite{clarkson:1936} modulus of convexity
\begin{equation}\label{eq:clarkson}
  \delta_{U}(\epsilon)
  :=
  \inf_{\substack{u,v\in S_{U}\\\left\|u-v\right\|_{U}=\epsilon}}
  \left(
    1
    -
    \left\|
      \frac{u+v}{2}
    \right\|_{U}
  \right),
  \quad
  \epsilon\in(0,2]
\end{equation}
(we will be mostly interested in the small values of $\epsilon$).

Let us say that a Banach space $\FFF$ of real-valued functions $f$ on $\mathbf{X}$
(with the standard pointwise operations of addition and scalar multiplication)
is a \emph{proper Banach functional space} (PBFS) on $\mathbf{X}$
if, for each $x\in\mathbf{X}$,
the evaluation functional $\kkk_x:f\in\FFF\mapsto f(x)$ is continuous.
We will assume that
\begin{equation}\label{eq:finiteness}
  \ccc_{\FFF}
  :=
  \sup_{x\in\mathbf{X}}
  \left\|
    \kkk_x
  \right\|_{\FFF^*}
  <
  \infty,
\end{equation}
where $\FFF^*$ is the dual Banach space
(see, e.g., \cite{rudin:1991}, Chapter 4).

The following theorem will be proved in \S\S\ref{sec:geometry}--\ref{sec:proof}.
\begin{theorem}\label{thm:main}
  Let $\FFF$ be a proper Banach functional space
  such that
  \begin{equation}\label{eq:condition}
    \forall\epsilon\in(0,2]:
    \delta_{\FFF}(\epsilon)\ge(\epsilon/2)^p/p
  \end{equation}
  for some $p\in[2,\infty)$.
  There exists a prediction algorithm producing $\mu_n\in[-Y,Y]$
  that are guaranteed to satisfy
  \begin{equation}\label{eq:main}
    \frac1N
    \sum_{n=1}^N
    \left(
      y_n-\mu_n
    \right)^2
    \le
    \frac1N
    \sum_{n=1}^N
    \left(
      y_n-D(x_n)
    \right)^2
    +
    40
    Y
    \sqrt{\ccc_{\FFF}^2+1}
    \left(
      \left\|
        D
      \right\|_{\FFF}
      +
      Y
    \right)
    N^{-1/p}
  \end{equation}
  for all $N=1,2,\ldots$ and all $D\in\FFF$.
\end{theorem}
Conditions (\ref{eq:finiteness}) and (\ref{eq:condition}) are satisfied
for the Sobolev spaces $W^{s,p}(\mathbf{X})$,
which we will now define.

\subsection*{Sobolev spaces}

Suppose $\mathbf{X}$ is an open or closed set in $\bbbr^m$.
(The standard theory assumes that $\mathbf{X}$ is open,
but the results we need easily extend to closed $\mathbf{X}$.)
We only define the Sobolev spaces $W^{s,p}(\mathbf{X})$
for the cases $s\in(0,1)$ and $p>m/s$;
for a more general definition
see, e.g., \cite{\Nikolsky} (pp.~57, 61)\ifnotCONF\ or
\cite{adams:1975} (Theorem 7.48 and Remark 7.49)\fi.

Let $s\in(0,1)$ and $p>m/s$.
For a function $f\in L^p(\mathbf{X})$ define
\begin{equation}\label{eq:norm}
  \left\|
    f
  \right\|_{s,p}
  :=
  \left(
    \int_{\mathbf{X}}
    \left|
      f(x)
    \right|^p
    \,\D x
    +
    \int_{\mathbf{X}}
    \int_{\mathbf{X}}
    \left|
      \frac{f(x)-f(y)}{\left|x-y\right|^s}
    \right|^p
    \frac{\D x \D y}{\left|x-y\right|^m}
  \right)^{1/p}
\end{equation}
(we use $\left|\cdot\right|$ to denote the Euclidean norm in $\bbbr^m$).
The Sobolev space $W^{s,p}(\mathbf{X})$ is defined to be the set of all $f$
such that $\left\|f\right\|_{s,p}<\infty$.
The Sobolev imbedding theorem says that,
for a wide range of $\mathbf{X}$
(definitely including our main example $\mathbf{X}=[0,1]\subseteq\bbbr$),
the functions in $W^{s,p}(\mathbf{X})$ can be made continuous
by a change on a set of measure zero;
we will always assume that this is true for our object space $\mathbf{X}$
and consider the elements of $W^{s,p}(\mathbf{X})$ to be continuous functions.
Let $C(\mathbf{X})$ be the Banach space of continuous functions
$f:\mathbf{X}\to\bbbr$
with finite norm
$
  \left\|
    f
  \right\|_{C(\mathbf{X})}
  :=
  \sup_{x\in\mathbf{X}}
  \left|
    f(x)
  \right|
$.
The Sobolev imbedding theorem also says that the imbedding
$W^{s,p}(\mathbf{X})\hookrightarrow C(\mathbf{X})$
(i.e., the function that maps each $f\in W^{s,p}(\mathbf{X})$
to the same function but considered as an element of $C(\mathbf{X})$)
is continuous,
i.e., that
\begin{equation*}
  \ccc_{s,p}:=\ccc_{W^{s,p}(\mathbf{X})}<\infty:
\end{equation*}
notice that $\ccc_{s,p}$ is just the norm of the imbedding
$W^{s,p}(\mathbf{X})\hookrightarrow C(\mathbf{X})$.
These conclusions depend on the condition $p>m/s$
(there are other parts of the Sobolev imbedding theorem,
dealing with the case where this condition is not satisfied).
For a proof in the case $\mathbf{X}=\bbbr^m$,
see, e.g., \cite{\AdamsFournier}, Theorems 7.34(c) and 7.47(a,c);
this implies the analogous statement\label{p:ext} for $\mathbf{X}$
with smooth boundary
since for such $\mathbf{X}$ every $f\in W^{s,p}(\mathbf{X})$
can be extended to an element of $W^{s,p}(\bbbr^m)$
without increasing the norm more than a constant times
(see, e.g., \cite{\Nikolsky}, p.~81).
We will say ``domain'' to mean a subset of $\bbbr^n$
which satisfies the conditions of regularity
mentioned in this paragraph.

The norm (\ref{eq:norm}) (sometimes called the Sobolev--Slobodetsky norm)
is only one of the standard norms giving rise
to the same topological vector space,
and the term ``Sobolev space'' is usually used to refer to the topology
rather than a specific norm;
in this paper we will not consider any other norms.
The restriction $s\in(0,1)$ is not essential for the results in this paper,
but the definition of $\left\|\cdot\right\|_{s,p}$
becomes slightly more complicated when $s\ge1$
(cf.\ \cite{\Nikolsky});
\cite{\AdamsFournier} gives a different but equivalent norm.

\ifnotCONF
For comparison purposes we will also define the spaces $W^{1,p}([0,1])$,
$p\in(1,\infty)$:
set
\begin{equation*}
  \left\|
    f
  \right\|_{1,p}
  :=
  \left(
    \int_0^1
    \left|
      f(x)
    \right|^p
    \,\D x
    +
    \int_0^1
    \left|
      f'(x)
    \right|^p
    \,\D x
  \right)^{1/p}
\end{equation*}
and include in $W^{1,p}([0,1])$
all absolutely continuous functions $f:[0,1]\to\bbbr$
with $\left\|f\right\|_{1,p}<\infty$.
We will always assume $\mathbf{X}=[0,1]$ in the case $s=1$.
\fi

We can now deduce the following corollary from Theorem~\ref{thm:main}.
It is known that (\ref{eq:condition}) is satisfied for the Sobolev spaces
$W^{s,p}(\mathbf{X})$
(see (\ref{eq:delta-Sobolev2})).
Let $p\in[2,\infty)$ and $s\in(m/p,1)$.
There exists a constant $C_{s,p}>0$
and a prediction algorithm producing $\mu_n\in[-Y,Y]$
that are guaranteed to satisfy
\begin{equation}\label{eq:main-Sobolev}
  \frac1N
  \sum_{n=1}^N
  \left(
    y_n-\mu_n
  \right)^2
  \le
  \frac1N
  \sum_{n=1}^N
  \left(
    y_n-D(x_n)
  \right)^2
  +
  Y
  C_{s,p}
  \left(
    \left\|
      D
    \right\|_{s,p}
    +
    Y
  \right)
  N^{-1/p}
\end{equation}
for all $N=1,2,\ldots$ and all $D\in W^{s,p}(\mathbf{X})$.

\ifnotCONF
In informal discussions below
we will continue to call terms such as the second addend
on the right-hand side of (\ref{eq:main-Sobolev})
the ``regret term'',
and say that the corresponding prediction algorithm
is ``$R$-competitive'',
where $R$ is the regret term. 
\fi

According to (\ref{eq:main}),
we can take
\begin{equation*}
  C_{s,p}
  =
  40
  \sqrt{\ccc_{s,p}^2+1},
\end{equation*}
but in fact
\begin{equation}\label{eq:C1}
  C_{s,p}
  =
  4
  \times
  8.68^{1-1/p}
  \sqrt{\ccc_{s,p}^2+1}
\end{equation}
will suffice (see (\ref{eq:C}) below).
In the special case $p=2$ one can use Hilbert-space methods
to improve (\ref{eq:C1}),
which now becomes, approximately,
\begin{equation}\label{eq:C2}
  11.78
  \sqrt{\ccc_{s,2}^2+1},
\end{equation}
to
\begin{equation}\label{eq:C3}
  2
  \sqrt{\ccc_{s,2}^2+1}
\end{equation}
(\cite{\GTPXI}, Theorem 1);
using Banach-space methods
we have lost a factor of $5.89$.
\ifnotCONF
For example, in the case $s=1$,
(\ref{eq:C2}) gives $C_{s,p}\approx 17.92$
and (\ref{eq:C3}) gives $C_{s,p}\approx 3.04$
(the value $\ccc_{1,2}^2=\coth1$ was found in \cite{marti:1983};
for further details of the case $s=1,p=2$, see \cite{\GTPXI}, \S4).
\fi

\subsection*{Application to the H\"older-continuous functions}

An important limiting case of the norm (\ref{eq:norm}) is
\begin{equation*}
  \left\|
    f
  \right\|_{s,\infty}
  :=
  \max
  \left(
    \sup_{x\in\mathbf{X}}
    \left|
      f(x)
    \right|,
    \sup_{x,y\in\mathbf{X}:x\ne y}
    \left|
      \frac{f(x)-f(y)}{\left|x-y\right|^s}
    \right|
  \right),
\end{equation*}
where $f:\mathbf{X}\to\bbbr$ is, as usual, assumed continuous.
The space $W^{s,\infty}(\mathbf{X})$ consists of the functions $f$
with $\left\|f\right\|_{s,\infty}<\infty$,
and its elements are called \emph{H\"older continuous of order $s$}.

The H\"older-continuous functions of order $s$
are perhaps the most intuitive formalization of
the functions with H\"older exponent $h\ge s$.
Let us see what Theorem \ref{thm:main} gives for them.

Suppose that $\mathbf{X}$ is a bounded domain in $\bbbr^m$,
$p\in(1,\infty)$,
and $s,s'\in(0,1)$ are such that $s'<s$.
If $f\in W^{s,\infty}(\mathbf{X})$,
\ifnotCONF
\begin{multline}\label{eq:imbedding}
  \left\|
    f
  \right\|_{s',p}
  =
  \left(
    \int_{\mathbf{X}}
    \left|
      f(x)
    \right|^p
    \,\D x
    +
    \int_{\mathbf{X}}
    \int_{\mathbf{X}}
    \left|
      \frac{f(x)-f(y)}{\left|x-y\right|^{s'}}
    \right|^p
    \frac{\D x \D y}{\left|x-y\right|^m}
  \right)^{1/p}\\
  \le
  \left(
    C^p
    +
    \int_{\mathbf{X}}
    \int_{\mathbf{X}}
    \left|
      \frac{C\left|x-y\right|^s}{\left|x-y\right|^{s'}}
    \right|^p
    \frac{\D x \D y}{\left|x-y\right|^m}
  \right)^{1/p}\\
  =
  \left(
    C^p
    +
    C^p
    \int_{\mathbf{X}}
    \int_{\mathbf{X}}
    \left|x-y\right|^{-m+sp-s'p}
    \,\D x \D y
  \right)^{1/p}\\
  \le
  \left(
    C^p
    +
    C^p
    \left|\mathbf{X}\right|
    \int_0^{\diam\mathbf{X}}
    t^{-m+sp-s'p}
    \frac{d}{dt}
    \left(
      \frac{\pi^{m/2}}{\Gamma(m/2+1)}
      t^m
    \right)
    \,\D t
  \right)^{1/p}\\
  =
  C
  \left(
    1
    +
    m
    \frac{\pi^{m/2}}{\Gamma(m/2+1)}
    \left|\mathbf{X}\right|
    \frac{(\diam\mathbf{X})^{(s-s')p}}{(s-s')p}
  \right)^{1/p},
\end{multline}
where $C:=\left\|f\right\|_{s,\infty}$,
\fi
$\left|\mathbf{X}\right|$ stands for the volume (Lebesgue measure) of $\mathbf{X}$\ifnotCONF,\fi
\ and $\diam\mathbf{X}$ stands for the diameter of $\mathbf{X}$
\ifnotCONF; remember that $\pi^{m/2}/\Gamma(m/2+1)$ is the volume of the unit ball in $\bbbr^m$. \fi
Therefore, (\ref{eq:imbedding})
gives an explicit bound for the norm of the continuous imbedding
$W^{s,\infty}(\mathbf{X})\hookrightarrow W^{s',p}(\mathbf{X})$.

Fix an arbitrarily small $\epsilon>0$.
Applying (\ref{eq:main-Sobolev}) to $W^{s',p}(\mathbf{X})$
with $p>m/s$ sufficiently close to $m/s$ and to $s'\in(m/p,s)$,
we can see from (\ref{eq:imbedding})
that there exists a constant $C_{s,\epsilon}>0$
such that
\begin{equation}\label{eq:main-Holder}
  \frac1N
  \sum_{n=1}^N
  \left(
    y_n-\mu_n
  \right)^2
  \le
  \frac1N
  \sum_{n=1}^N
  \left(
    y_n-D(x_n)
  \right)^2
  +
  Y
  C_{s,\epsilon}
  \left(
    \left\|
      D
    \right\|_{s,\infty}
    +
    Y
  \right)
  N^{-s/m+\epsilon}
\end{equation}
holds for all $N=1,2,\ldots$ and all $D\in W^{s,\infty}(\mathbf{X})$.

\section{Implications for a stochastic Reality}
\label{sec:stochastic}

In this section we discuss implications
of Theorem \ref{thm:main} for statistical learning theory
and filtering of random processes.
Surprisingly,
even when Reality follows a specific stochastic strategy,
competitive on-line results do not trivialize
but provide new meaningful information.

\subsection*{Statistical learning theory}

In this section
we apply the method of \cite{cesabianchi/etal:2004}
to derive a corollary of Theorem \ref{thm:main}
for the statistical learning framework,
where $(x_n,y_n)$ are assumed to be drawn independently
from the same probability distribution on $\mathbf{X}\times[-Y,Y]$.

The \emph{risk} of a prediction rule
(formally, a measurable function) $D:\mathbf{X}\to\bbbr$
with respect to a probability distribution $P$ on $\mathbf{X}\times[-Y,Y]$
is defined as
\begin{equation*}
  \risk_P(D)
  :=
  \int_{\mathbf{X}\times[-Y,Y]}
    (y-D(x))^2
  P(\dd x,\dd y).
\end{equation*}
Our current goal is to construct,
from a given sample,
a prediction rule whose risk
is competitive with the risk of small-norm prediction rules
in $W^{s,p}(\mathbf{X})$.

Fix an on-line prediction algorithm
and a sequence $(x_1,y_1),(x_2,y_2),\ldots$ of examples.
For each $n=1,2,\ldots$ and each $x\in\mathbf{X}$,
define $H_n(x)$ to be the prediction $\mu_n\in\bbbr$
output by the algorithm
when fed with $(x_1,y_1),\ldots,(x_{n-1},y_{n-1}),x$.
We will assume that the functions $H_n$ are always measurable
(they are for our algorithm\ifnotCONF, constructed in the following two sections\fi).
The prediction rule
\begin{equation*}
  \overline{H}_N(x)
  :=
  \frac1N
  \sum_{n=1}^N
  H_n(x)
\end{equation*}
will be said to be \emph{obtained by averaging} from the prediction algorithm.

The following result is an easy application of the method of \cite{cesabianchi/etal:2004}
to (\ref{eq:main-Sobolev});
we refrain from stating the analogous result based on (\ref{eq:main-Holder}).
\begin{corollary}\label{cor:main1}
  Let $\mathbf{X}$ be a domain in $\bbbr^m$, $p\ge2$, $s\in(m/p,1)$,
  and let $\overline{H}_N$, $N=1,2,\ldots$,
  be the prediction rule obtained by averaging
  from some prediction algorithm guaranteeing (\ref{eq:main-Sobolev}).
  For any $D\in W^{s,p}(\mathbf{X})$,
  any probability distribution $P$ on $\mathbf{X}\times[-Y,Y]$,
  any $N=1,2,\ldots$,
  and any $\delta>0$,
  \begin{equation}\label{eq:cor1}
    \risk_P(\overline{H}_N)
    \le
    \risk_P(D)
    +
    Y
    C_{s,p}
    \left(
      \left\|
        D
      \right\|_{s,p}
      +
      Y
    \right)
    N^{-1/p}
    +
    4Y^2
    \sqrt{2\ln\frac{2}{\delta}}
    N^{-1/2}
  \end{equation}
  with probability at least $1-\delta$.
\end{corollary}
\ifnotCONF
\begin{proof}
  Without loss of generality we assume that $D(x)\in[-Y,Y]$ for all $x\in\mathbf{X}$
  and that $H_n(x)\in[-Y,Y]$ for all $x\in\mathbf{X}$ and $n$.
  Outside an event of probability
  \begin{equation}\label{eq:delta}
     \delta
     :=
     2
     \exp
     \left(
       -\frac{\epsilon^2 N}{8Y^4}
     \right)
  \end{equation}
  we have
  (some steps will be explained later on)
  \begin{align}
    \risk_P(\overline{H}_N)
    &\le
    \frac1N
    \sum_{n=1}^N
    \risk_P(H_n)
    \label{eq:jensen}\\
    &\le
    \frac1N
    \sum_{n=1}^N
    (y_n-H_n(x_n))^2
    +
    \epsilon
    \label{eq:hoeffding1}\\
    &\le
    \frac1N
    \sum_{n=1}^N
    (y_n-D(x_n))^2
    +
    Y
    C_{s,p}
    \left(
      \left\|D\right\|_{s,p} + Y
    \right)
    N^{-1/p}
    +
    \epsilon
    \label{eq:I}\\
    &\le
    \frac1N
    \sum_{n=1}^N
    \risk_P(D)
    +
    Y
    C_{s,p}
    \left(
      \left\|D\right\|_{s,p} + Y
    \right)
    N^{-1/p}
    +
    2\epsilon
    \label{eq:hoeffding2}\\
    &=
    \risk_P(D)
    +
    Y
    C_{s,p}
    \left(
      \left\|D\right\|_{s,p} + Y
    \right)
    N^{-1/p}
    +
    2\epsilon.
    \label{eq:last}
  \end{align}
  The first inequality, (\ref{eq:jensen}),
  follows from the convexity of the function $t\mapsto t^2$.
  Inequalities (\ref{eq:hoeffding1}) and (\ref{eq:hoeffding2})
  follow from Hoeffding's martingale inequality
  (\cite{hoeffding:1963}; see also \cite{devroye/etal:1996}, Theorem 9.1 on p.~135).
  Either of (\ref{eq:hoeffding1}) and (\ref{eq:hoeffding2})
  holds with probability at least $1-\delta/2$;
  therefore, both will hold with probability at least $1-\delta$.
  Finally, inequality (\ref{eq:I}) follows from (\ref{eq:main-Sobolev}).

  Our goal, (\ref{eq:cor1}),
  follows from the inequality between the extreme terms
  of (\ref{eq:jensen})--(\ref{eq:last})
  if we substitute
  \begin{equation}\label{eq:epsilon}
    \epsilon
    =
    2Y^2
    \sqrt{2\ln\frac{2}{\delta}}
    N^{-1/2}
  \end{equation}
  (which is a different way of writing (\ref{eq:delta})).
  \qedtext
\end{proof}

For a fixed $\delta$, the regret term
(the sum of the second and third addends on the right-hand side)
of (\ref{eq:cor1}) grows as $N^{-1/p}$.
For a discussion of related results in statistical learning theory,
see \cite{\GTPXI} (versions 1 and 2), \S5.
\fi

\subsection*{Filtering of random processes}

Suppose we are interested in the value of a ``signal'' $\Theta:[0,1]\to\bbbr$
sequentially observed at moments $t_n:=n/N$, $n=1,\ldots,N$,
where $N$ is a large positive integer;
let $\theta_n:=\Theta(t_n)$.
The problem is that our observations of $\theta_n$ are imperfect,
and in fact we see $y_n=\theta_n+\xi_n$,
where each noise random variable $\xi_n$ has zero expectation
given the past.
We assume that $\Theta$ belongs to $W^{s,p}([0,1])$
(but do not make any assumptions about the mechanism,
deterministic, stochastic, or other, that generated it)
and that $\theta_n,y_n\in[-Y,Y]$ for a known constant $Y$.
Let us use the $\mu_n$ from Theorem~\ref{thm:main}
as estimates of the true values $\theta_n$.
The elementary equality
\begin{equation}\label{eq:elementary}
  a^2
  =
  (a-b)^2-b^2+2ab
\end{equation}
implies
\begin{equation}\label{eq:fundamental}
  \sum_{n=1}^N
  (\mu_n-\theta_n)^2
  =
  \sum_{n=1}^N
  (y_n-\mu_n)^2
  -
  \sum_{n=1}^N
  (y_n-\theta_n)^2
  +
  2
  \sum_{n=1}^N
  (y_n-\theta_n)
  (\mu_n-\theta_n).
\end{equation}
Hoeffding's inequality in the martingale form shows that, for any $C>0$,
\begin{equation*}
  \Prob
  \left\{
    2
    \sum_{n=1}^N
    (y_n-\theta_n)
    (\mu_n-\theta_n)
    \ge
    C
  \right\}
  \le
  \exp
  \left(
    -\frac{C^2}{128Y^4N}
  \right).
\end{equation*}
Substituting this (with $C$ expressed via the right-hand side, denoted $\delta$)
and (\ref{eq:main-Sobolev}) into (\ref{eq:fundamental}),
we obtain the following corollary,
which we state somewhat informally.
\begin{corollary}\label{cor:filtering}
  Let $p\ge2$, $s\in(1/p,1)$, and $\delta>0$.
  Suppose that $\Theta\in W^{s,p}([0,1])$ and $y_n=\theta_n+\xi_n\in[-Y,Y]$,
  where $\theta_n:=\Theta(n/N)\in[-Y,Y]$
  and $\xi_n$ are random variables whose expectation given the past (including $\theta_n$) is zero.
  With probability at least $1-\delta$
  the $\mu_n$ of (\ref{eq:main-Sobolev}) satisfy
  \begin{equation}\label{eq:cor-filtering}
    \frac1N
    \sum_{n=1}^N
    \left(
      \mu_n-\theta_n
    \right)^2
    \le
    Y
    C_{s,p}
    \left(
      \left\|
        \Theta
      \right\|_{s,p}
      +
      Y
    \right)
    N^{-1/p}
    +
    8Y^2
    \sqrt{2\ln\frac{1}{\delta}}
    N^{-1/2}.
  \end{equation}
\end{corollary}
The constant $C_{s,p}$ in (\ref{eq:cor-filtering}) is the one in (\ref{eq:C1}).
From (\ref{eq:main-Holder}),
we can also see that, if we assume $\Theta\in W^{s,\infty}([0,1])$,
\begin{equation}\label{eq:filtering-Holder}
  \frac1N
  \sum_{n=1}^N
  \left(
    \mu_n-\theta_n
  \right)^2
  \le
  Y
  C_{s,\epsilon}
  \left(
    \left\|
      \Theta
    \right\|_{s,\infty}
    +
    Y
  \right)
  N^{-s+\epsilon}
  +
  8Y^2
  \sqrt{2\ln\frac{1}{\delta}}
  N^{-1/2}
\end{equation}
will hold with probability at least $1-\delta$.

It is important that the function $\Theta$
in (\ref{eq:cor-filtering}) and (\ref{eq:filtering-Holder})
does not have to be chosen in advance:
it can be constructed ``step-wise'',
with $\Theta(t)$ for $t\in(n/N,(n+1)/N]$ chosen at will after observing $\xi_n$
and taking into account all other information that becomes available
before and including time $n/N$.
A clean formalization of this intuitive picture
seems to require the game-theoretic probability of \cite{shafer/vovk:2001}
(although we can get the picture ``almost right''
using the standard measure-theoretic probability).

In the case where $\Theta$ is generated from a diffusion process,
it will almost surely belong to $W^{(1-\epsilon)/2,\infty}([0,1])$
(this follows from standard results about the Brownian motion,
such as L\'evy's modulus theorem:
see, e.g., \cite{karatzas/shreve:1991}, Theorem 9.25),
and so the regret term in (\ref{eq:cor-filtering}) and (\ref{eq:filtering-Holder})
can be made $O(N^{-1/2+\epsilon})$,
for an arbitrarily small $\epsilon>0$.
The Kalman filter,
which is stochastically optimal,
gives a somewhat better regret, $O(N^{-1/2})$.
Corollary \ref{cor:filtering}, however, does not depend on the very specific assumptions of
the Kalman filter:
we do not require the linearity, Gaussianity, or even stochasticity of the model;
the assumption about the noise $\xi_n$ is minimal (zero expectation given the past).
Instead, we have the assumption that all $\theta_n$ and $y_n$ are chosen from $[-Y,Y]$.
It appears that in practice the interval to which the $\theta_n$ and $y_n$ are assumed to belong
should change slowly as new data are processed.
This is analogous to the situation with the Kalman filter,
which, despite assuming linear systems, has found its greatest application
to non-linear systems \cite{sorenson:1970};
what is usually used in practice is the ``extended Kalman filter'',
which relies on a slowly changing linearization of the non-linear system.

\ifnotCONF
Until the end of this section
we will discuss in more detail
the standard stochastic approach to the problem of filtering
(\cite{kalman:1960}; see also \cite{sorenson:1970}, \cite{shiryaev:1996}, \S VI.7,
and, for a continuous-time version,
\cite{kalman/bucy:1961}, \cite{\LiptserShiryaev}, \S10.1).
The signal is now modeled as a random process $\Theta_t$, $t\in[0,1]$,
governed by the stochastic differential equation
\begin{equation}\label{eq:Theta}
  d\Theta_t
  =
  \left(
    a_0(t)
    +
    a_1(t) \Theta_t
  \right)
  dt
  +
  b(t) dB_t,
\end{equation}
where $B_t$ is the standard Brownian motion
(a zero-mean Gaussian continuous stochastic process on $[0,1]$
such that $B_0=0$ and the variance of each increment $B_{t_1}-B_{t_2}$
is $\left|t_1-t_2\right|$)
and $a_0,a_1,b:[0,1]\to\bbbr$ are bounded Borel functions.
The process starts from a random value $\Theta_0$
(modeled as a Gaussian random variable independent of $B_t$)
and, as before, is observed at points $t_n:=n/N$;
$\theta_n:=\Theta(t_n)$.
The observed sequence
is $y_n=\theta_n+\sigma\xi_n$
(neither $\theta_n$ nor $y_n$ are assumed to be bounded by a known constant),
where $\sigma$ is a positive constant
and $\xi_n$ are standard Gaussian random variables
independent between themselves
and of the initial position $\Theta_0$ and the Brownian motion $B_t$.
In some important respects this is a simplification of the usual filtering problems;
e.g., we consider scalar rather than vector $\Theta_t$ and $y_n$.

Earlier we discussed the possibility of positive contributions
of competitive on-line results,
such as Theorem \ref{thm:main},
to the problem of filtering,
and now we will briefly explore the connection in the opposite direction:
limitations on competitive on-line prediction
following from the known optimality properties of the Kalman filter.
According to (\ref{eq:main-Holder}),
there is a prediction algorithm $O(N^{-s+\epsilon})$-competitive
with $W^{s,\infty}([0,1])$,
for any $\epsilon>0$.
It remains an open problem to show that the rate $N^{-s+\epsilon}$
(we will disregard plus or minus $\epsilon$ in the rest of this section)
cannot be improved,
but the following considerations make it likely in the case $s\approx1/2$.
(For an alternative argument,
see, e.g., Theorem 4 in \cite{\GTPXI}.)

Suppose the prediction rule $D:[0,1]\to\bbbr$ is generated randomly
as the trajectory of the stochastic process (\ref{eq:Theta})
with $\Theta_0=0$, $a_0(t)\equiv0$, $a_1(t)\equiv0$,
and $b(t)\equiv c>0$
(i.e., $D(t)=c B_t$, where $B$ is the standard Brownian motion).
The positive constant $c$ is chosen small as compared to $Y$,
so that $D(t)$ is unlikely to take values approaching $-Y$ or $Y$.
It is clear that the observations $y_n$ are generated independently
(given $B$)
from the normal distribution $N(D(t_n),\sigma^2)$
with mean $D(t_n)$ and variance $\sigma^2$;
if $y_n$ falls outside $[-Y,Y]$, it is truncated to $Y\sign y_n$.
The variance $\sigma^2>0$ is assumed to be small enough
for the probability of $\left|y_n\right|<Y$ to be close to 1 for each $n$
(or we can even take $c$ and $\sigma$ slightly, say logarithmically, dependent on $N$
so that $\max_n\left|y_n\right|<Y$ with a probability tending to 1).
According to the standard properties of the Kalman filter
(see, e.g., \cite{\LiptserShiryaev}, Theorem 13.4,
or \cite{shiryaev:1996}, Theorem VI.7.1),
the variance $\gamma_n$ of the best estimate of $\theta_n$
(which is also the best estimate of $y_n$), $n>1$,
given $y_1,\ldots,y_{n-1}$ satisfies the recurrent equation
\begin{equation*}
  \gamma_{n+1}
  =
  \gamma_n
  +
  \frac{c^2}{N}
  -
  \frac{\gamma_n^2}{\sigma^2+\gamma_n}.
\end{equation*}
It is clear that $\gamma_n$ is an increasing sequence
tending, as $n\to\infty$, to a limit equal to
\begin{equation*}
  \frac{c^2+\sqrt{c^4+4c^2\sigma^2N}}{2N}
  >
  \frac{c\sigma}{\sqrt{N}},
\end{equation*}
and that it will move significantly towards this limit
already during the first $\surd N$ rounds
(cf.\ Figure \ref{fig:kalman}).
By Hoeffding's inequality,
the excess of the total loss of the stochastically best algorithm
(the Kalman filter)
over the total loss of $D$ will be of order $N^{1/2}$,
and so the excess of its average loss will be of order $N^{-1/2}$
(with probability very close to $1$).

Since the sample paths of diffusion processes
almost surely belong to $W^{s,\infty}([0,1])$
for all $s\in(0,1/2)$,
we can see that no prediction algorithm can be $O(N^{-1/2-\epsilon})$-competitive
with $W^{1/2-\epsilon,\infty}([0,1])$.
Therefore, if we disregard the epsilons,
our algorithm achieves the optimal rate of decay in $N$
of the regret term for $s\approx1/2$.

\begin{figure}[bt]
  \centering
  \makebox{\includegraphics[width=10cm,height=5cm]{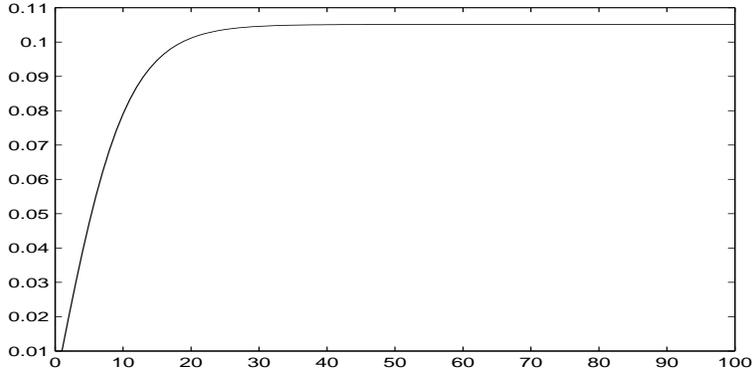}}
  \caption{\label{fig:kalman}The growth of the Kalman filter's error $\gamma_n$, $n=1,\ldots,N$,
    for $c=\sigma^2=1$ and $N=100$; the final value $\gamma_N$ is approximately $N^{-1/2}$.}
\end{figure}

A similar argument might have also worked in the case $s<1/2$
had we known an analogue of the Kalman filter result
for the fractional Brownian motion,
where $B$ is replaced with a stochastic process $B^{(h)}$, $h\in(0,1/2)$,
defined in the same way except that
the variance of each increment $B^{(h)}_{t_1}-B^{(h)}_{t_2}$
is $\left|t_1-t_2\right|^{2h}$
(notice that $B=B^{(1/2)}$).
Unfortunately, we know of no such result,
although a step in this direction is made in \cite{nuzman/poor:2000}.
\fi

\section{More geometry of Banach spaces}
\label{sec:geometry}

\ifnotCONF
In the proof of Theorem \ref{thm:main}
we will need not only Clarkson's modulus of convexity (\ref{eq:clarkson})
but a whole range of different moduli of convexity and smoothness.
In our description we will often follow \cite{lindenstrauss/tzafriri:1979};
for information about other moduli and further references,
see \cite{fuster:2005}.
We will only consider Banach spaces of dimension at least $2$.

\subsection*{Moduli of convexity and smoothness}

A natural modification of Clarkson's modulus of convexity
was proposed by Gurary \cite{\Gurary}:
\begin{equation}\label{eq:gurary}
  \delta^{\dagger}_{U}(\epsilon)
  :=
  \inf_{\substack{u,v\in S_{U}\\\left\|u-v\right\|_{U}=\epsilon}}
  \left(
    1
    -
    \inf_{t\in[0,1]}
    \left\|
      tu + (1-t)v
    \right\|_{U}
  \right).
\end{equation}
It is clear that
\begin{equation*}
  \delta_{U}(\epsilon) \le \delta^{\dagger}_{U}(\epsilon) \le 2\delta_{U}(\epsilon)
\end{equation*}
(cf.\ the proof of Lemma \ref{lem:rho-1-2} below),
and it was shown recently \cite{barcenas/etal:2004} that this relation cannot be improved.
\fi

The standard modulus of smoothness 
was proposed by Lindenstrauss \cite{lindenstrauss:1963}:
\begin{equation}\label{eq:lindenstrauss}
  \rho_U(\tau)
  :=
  \sup_{u,v\in S_{U}}
  \left(
    \frac
    {
      \left\|
        u + \tau v
      \right\|_{U}
      +
      \left\|
        u - \tau v
      \right\|_{U}
    }
    {2}
    -
    1
  \right),
  \quad
  \tau>0.
\end{equation}
Lindenstrauss also established a simple but very useful relation
of conjugacy
(cf.\ \cite{rockafellar:1970}, \S12,
although $\delta$ is not always convex \cite{\Liokumovich})
between $\delta$ and $\rho$:
\begin{equation}\label{eq:fenchel}
  \rho_{U^*}(\tau)
  =
  \sup_{\epsilon\in(0,2]}
  \left(
    \frac{\epsilon\tau}{2}
    -
    \delta_{U}(\epsilon)
  \right);
\end{equation}
we can see that $2\rho_{U^*}$ is the Fenchel transform of $2\delta_U$.

The following inequality will be the basis of the proof of Theorem \ref{thm:main}
in the next section.
Suppose a PBFS $\FFF$ satisfies the condition (\ref{eq:condition}) of Theorem \ref{thm:main}.
By (\ref{eq:fenchel}) we obtain for the dual space $\FFF^*$ to $\FFF$,
assuming $\tau\in(0,1]$:
\begin{equation}\label{eq:dual}
  \rho_{\FFF^*}(\tau)
  \le
  \sup_{\epsilon\in(0,2]}
  \left(
    \frac{\epsilon\tau}{2}
    -
    (\epsilon/2)^p/p
  \right)
  =
  \tau^q/q,
\end{equation}
where $q:=p/(p-1)$
(the supremum in (\ref{eq:dual}) is attained
at $\epsilon=2\tau^{1/(p-1)}$).

\ifnotCONF
The Banach space $U$ is called \emph{uniformly convex}
if $\delta_{U}(\epsilon)>0$ for all $\epsilon\in(0,2]$,
and it is called \emph{uniformly smooth}
if $\rho_{U}(\tau)\to0$ as $\tau\to0$.
All uniformly convex and all uniformly smooth Banach spaces $U$ are reflexive
(i.e., $U^{**}=U$;
see, e.g., \cite{lindenstrauss/tzafriri:1979}, Proposition 1.e.3 on p.~61).
\fi

If $V$ is a Hilbert space,
the ``parallelogram identity''
\begin{equation}\label{eq:parallelogram}
  \left\|
    u+v
  \right\|_V^2
  +
  \left\|
    u-v
  \right\|_V^2
  =
  2
  \left\|
    u
  \right\|_V^2
  +
  2
  \left\|
    v
  \right\|_V^2
\end{equation}
immediately gives
\begin{equation*}
  \delta_{V}(\epsilon)
  =
  1
  -
  \sqrt{1-(\epsilon/2)^2}
  \ge
  \epsilon^2/8
\end{equation*}
and
\begin{equation}\label{eq:Hilbert}
  \rho_{V}(\tau)
  =
  \sqrt{1+\tau^2}
  -
  1
  \le
  \tau^2/2.
\end{equation}
\ifnotCONF
N\"ordlander \cite{nordlander:1960} proved
that the unit balls in Hilbert spaces are most convex and smooth:
if $U$ is a Banach space and $V$ is a Hilbert space,
\begin{equation}\label{eq:Nordlander}
  \begin{aligned}
    \delta_U(\epsilon)
    &\le
    \delta_V(\epsilon)
    =
    1
    -
    \sqrt{1-(\epsilon/2)^2},
  \\
    \rho_U(\tau)
    &\ge
    \rho_V(\tau)
    =
    \sqrt{1+\tau^2}
    -
    1.
  \end{aligned}
\end{equation}

The original definitions (\ref{eq:clarkson}) and (\ref{eq:lindenstrauss})
of the moduli of convexity and smoothness look very different,
and Bana\'s \cite{banas:1986} proposed a definition of modulus of smoothness
similar to (\ref{eq:clarkson}):
\begin{equation}\label{eq:banas}
  \rho^{\dagger}_{U}(\tau)
  :=
  \sup_{\substack{u,v\in S_{U}\\\left\|u-v\right\|_{U}=\tau}}
  \left(
    1
    -
    \left\|
      \frac{u+v}{2}
    \right\|_{U}
  \right),
  \quad
  \tau\in(0,2).
\end{equation}
The difference $\rho^{\dagger}_{U}(\epsilon)-\delta_{U}(\epsilon)$
measures the degree to which (the unit ball in) $U$ is deformed
\cite{banas/fraczek:1993}
(it is always zero for Hilbert spaces).
What we will need in this paper is the modification of (\ref{eq:banas})
in the direction of (\ref{eq:gurary}):
\begin{equation}\label{eq:my}
  \rho^{\ddagger}_{U}(\tau)
  :=
  \sup_{\substack{u,v\in S_{U}\\\left\|u-v\right\|_{U}=\tau}}
  \sup_{t\in[0,1]}
  \left(
    1
    -
    \left\|
      tu + (1-t)v
    \right\|_{U}
  \right),
  \quad
  \tau\in(0,2).
\end{equation}

Since the standard results about moduli of convexity and smoothness
are about the definitions (\ref{eq:clarkson}) and (\ref{eq:lindenstrauss}),
we first need to establish connections between (\ref{eq:lindenstrauss}) and (\ref{eq:my}).
The first of these results appears in \cite{banas:1986}
(but we still prove it since \cite{banas:1986} is less easily accessible
than most other papers in our bibliography).
\begin{lemma}[\cite{banas:1986}]\label{lem:rho-0-1}
  For all $\tau\in(0,2)$,
  \begin{equation}\label{eq:rho-0-1}
    \frac{\rho^{\dagger}_U(\tau)}{1-\rho^{\dagger}_U(\tau)}
    \le
    \rho_U
    \left(
      \frac
      {\tau}
      {
        2
        \left(
          1
          -
          \rho^{\dagger}_U(\tau)
        \right)
      }
    \right).
  \end{equation}
\end{lemma}

\begin{figure}[bt]
  \centering
  \makebox{\includegraphics[width=5cm,angle=270]{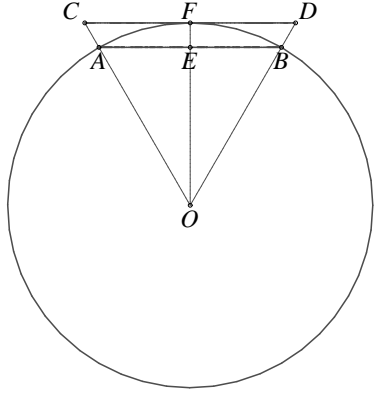}}
  \caption{\label{fig:smoothness}Relation between $\rho$ and $\rho^{\dagger}$.}
\end{figure}

\begin{proof}
  Let $c<\rho^{\dagger}_U(\tau)$ be such that,
  for some $u,v\in S_{U}$ satisfying $\left\|u-v\right\|_{U}=\tau$,
  \begin{equation*}
    \left\|
      \frac{u+v}{2}
    \right\|_{U}
    =
    1-c
  \end{equation*}
  (it is clear that $c$ can be chosen as close to $\rho^{\dagger}_U(\tau)$ as we wish).
  Set
  \begin{equation*}
    u'
    :=
    \frac{1}{1-c}
    \frac{u+v}{2},
    \quad
    v'
    :=
    \frac{v-u}{\left\|u-v\right\|_U},
    \quad
    \tau'
    :=
    \frac{1}{1-c}
    \frac{\tau}{2}
  \end{equation*}
  (cf.\ Figure \ref{fig:smoothness},
  where $\overrightarrow{OA}=u$,
  $\overrightarrow{OB}=v$,
  $\overrightarrow{OE}=(u+v)/2$,
  $\overrightarrow{OF}=u'$,
  and $\overrightarrow{FD}=\tau'v'$).
  Since $u',v'\in S_U$,
  we have
  \begin{equation*}
    \rho_U(\tau')
    \ge
    \frac
    {
      \left\|
        u' + \tau' v'
      \right\|_{U}
      +
      \left\|
        u' - \tau' v'
      \right\|_{U}
    }
    {2}
    -
    1
    =
    \frac{1}{1-c}-1,
  \end{equation*}
  which can be rewritten as
  \begin{equation*}
    \rho_U
    \left(
      \frac{\tau}{2(1-c)}
     \right)
    \ge
    \frac{c}{1-c}.
  \end{equation*}
  Letting $c\to\rho^{\dagger}_U(\tau)$ completes the proof
  (the modulus of smoothness is continuous by, e.g.,
  \cite{lindenstrauss/tzafriri:1979}, Proposition 1.e.5 on p.~64).
  \qedtext
\end{proof}

\begin{corollary}\label{cor:rho-0-1}
  For all $\tau\in(0,1]$,
  \begin{equation}\label{eq:cor-rho}
    \rho^{\dagger}_U(\tau)
    \le
    \rho_U(\tau).
  \end{equation}
\end{corollary}

\begin{proof}
  Let $\tau\in(0,1]$.
  Following \cite{banas:1986}, proof of Lemma 1, we obtain
  \begin{multline*}
    \rho^{\dagger}_U(\tau)
    =
    \sup_{\substack{u,v\in S_{U}\\\left\|u-v\right\|_{U}=\tau}}
    \frac
    {
      2
      \left\|
        u
      \right\|_{U}
      -
      \left\|
        u+v
      \right\|_{U}
    }
    {2}\\
    \le
    \sup_{\substack{u,v\in S_{U}\\\left\|u-v\right\|_{U}=\tau}}
    \frac
    {
      \left\|
        u+v
      \right\|_{U}
      +
      \left\|
        u-v
      \right\|_{U}
      -
      \left\|
        u+v
      \right\|_{U}
    }
    {2}
    =
    \frac{\tau}{2}
    \le
    \frac12.
  \end{multline*}
  We can now easily deduce (\ref{eq:cor-rho}) from (\ref{eq:rho-0-1})
  and the fact that $\rho_U$ is a non-decreasing function
  (\cite{lindenstrauss/tzafriri:1979}, Proposition 1.e.5):
  \begin{equation*}
    \rho^{\dagger}_U(\tau)
    \le
    \frac{\rho^{\dagger}_U(\tau)}{1-\rho^{\dagger}_U(\tau)}
    \le
    \rho_U
    \left(
      \frac
      {\tau}
      {
        2
        \left(
          1
          -
          \rho^{\dagger}_U(\tau)
        \right)
      }
    \right)
    \le
    \rho_U(\tau).
    \qedmath
  \end{equation*}
\end{proof}

\begin{lemma}\label{lem:rho-1-2}
  For all $\tau\in(0,2)$,
  \begin{equation*}
    \rho^{\ddagger}_U(\tau)
    \le
    2\rho^{\dagger}_U(\tau).
  \end{equation*}
\end{lemma}
\begin{proof}
  Suppose $\rho^{\ddagger}_U(\tau)>c$.
  Let $u,v\in S_U$ and $t\in[0,1]$ be such that $\left\|u-v\right\|_U=\tau$
  and
  \begin{equation*}
    \left\|
      t u + (1-t) v
    \right\|_U
    <
    1-c.
  \end{equation*}
  Without loss of generality we assume $t\le1/2$.
  Since
  \begin{multline*}
    \left\|
      \frac{u+v}{2}
    \right\|_U
    =
    \left\|
      \frac{1-2t}{2-2t}
      u
      +
      \frac{1}{2-2t}
      \left(
        t u + (1-t) v
      \right)
    \right\|_U\\
    \le
    \frac{1-2t}{2-2t}
    \left\|
      u
    \right\|_U
    +
    \frac{1}{2-2t}
    \left\|
      t u + (1-t) v
    \right\|_U
    <
    \frac{1-2t}{2-2t}
    +
    \frac{1}{2-2t}
    (1-c)\\
    =
    \frac{2-2t-c}{2-2t}
    \le
    \frac{2-c}{2}
    =
    1-\frac{c}{2},
  \end{multline*}
  we have $\rho^{\dagger}_U(\tau)>c/2$.
  \qedtext
\end{proof}

\subsection*{Direct sums of uniformly smooth spaces}
\fi

If $U_1$ and $U_2$ are two Banach spaces,
their \emph{weighted direct sum} $U_1\oplus U_2$
is defined to be the Cartesian product $U_1\times U_2$
with the operations of addition and multiplication by scalar
defined by
\begin{equation*}
  (u_1,u_2)
  +
  (u'_1,u'_2)
  :=
  (u_1+u'_1,u_2+u'_2),
  \quad
  c(u_1,u_2)
  :=
  (cu_1,cu_2);
\end{equation*}
we will equip it with the norm
\begin{equation}\label{eq:Euclidean}
  \left\|
    (u_1,u_2)
  \right\|_{U_1\oplus U_2}
  :=
  \sqrt
  {
    a_1
    \left\|
      u_1
    \right\|_{U_1}^2
    +
    a_2
    \left\|
      u_2
    \right\|_{U_2}^2
  },
\end{equation}
where $a_1$ and $a_2$ are positive constants
(to simplify formulas,
we do not mention them explicitly in our notation for $U_1\oplus U_2$).
The operation of weighted direct sum provides a means of merging different Banach spaces,
which plays an important role in our proof technique
(cf.\ \cite{\GTPXI}, Corollary 4).
The ``Euclidean'' definition (\ref{eq:Euclidean}) of the norm
in the direct sum suggests that the sum will be as smooth as the components;
this intuition is formalized in the following lemma
(essentially a special case of Proposition 17 in \cite{figiel:1976full}, p.~132).
\begin{lemma}\label{lem:direct}
  If $U_1$ and $U_2$ are Banach spaces
  and $f:(0,1]\to\bbbr$,
  \begin{multline*}
    \left(
      \forall\tau\in(0,1]:
      \rho_{U_1}(\tau)\le f(\tau)
      \;\&\;
      \rho_{U_2}(\tau)\le f(\tau)
    \right)\\
    \Longrightarrow
    \left(
      \forall\tau\in(0,1]:
      \rho_{U_1\oplus U_2}(\tau)\le 4.34 f(\tau)
    \right).
  \end{multline*}
\end{lemma}
\ifnotCONF
\begin{proof}
  We will follow the proof of Proposition 17 in \cite{figiel:1976full},
  which is based on the following weak form
  of the parallelogram identity (\ref{eq:parallelogram}),
  valid for all Banach spaces:
  \begin{multline}\label{eq:pseudo-parallelogram-a}
    \left\|
      u+v
    \right\|_U^2
    +
    \left\|
      u-v
    \right\|_U^2
    -
    2
    \left\|
      u
    \right\|_U^2
    -
    2
    \left\|
      v
    \right\|_U^2\\
    \le
    2
    \left\|
      u
    \right\|_U
    \left(
      \left\|
        u+v
      \right\|_U
      +
      \left\|
        u-v
      \right\|_U
      -
      2
      \left\|
        u
      \right\|_U
    \right)
  \end{multline}
  (see \cite{figiel:1976full}, Lemma 16 on p.~132);
  it is clear that (\ref{eq:pseudo-parallelogram-a}) implies
  \begin{equation}\label{eq:pseudo-parallelogram-b}
    \left\|
      u+v
    \right\|_U^2
    +
    \left\|
      u-v
    \right\|_U^2
    -
    2
    \left\|
      u
    \right\|_U^2
    -
    2
    \left\|
      v
    \right\|_U^2
    \le
    4
    \left\|
      u
    \right\|_U^2
    \rho_U
    \left(
      \left\|
        v
      \right\|_U
      /
      \left\|
        u
      \right\|_U
    \right).
  \end{equation}

  Let $u^{\dagger}=(u_1,u_2)$ and $v^{\dagger}=(v_1,v_2)$
  be arbitrary norm one vectors in $U_1\oplus U_2$.
  Applying (\ref{eq:pseudo-parallelogram-b})
  to $(u,v):=(u_1,\tau v_1)$ and $(u,v):=(u_2,\tau v_2)$,
  we obtain
  \begin{multline}\label{eq:pseudo-parallelogram1}
    \left\|
      u_1 + \tau v_1
    \right\|_{U_1}^2
    +
    \left\|
      u_1 - \tau v_1
    \right\|_{U_1}^2
    -
    2
    \left\|
      u_1
    \right\|_{U_1}^2
    -
    2 \tau^2
    \left\|
      v_1
    \right\|_{U_1}^2\\
    \le
    4
    \left\|
      u_1
    \right\|_{U_1}^2
    \rho_{U_1}
    \left(
      \tau
      \left\|
        v_1
      \right\|_{U_1}
      /
      \left\|
        u_1
      \right\|_{U_1}
    \right)
  \end{multline}
  and
  \begin{multline}\label{eq:pseudo-parallelogram2}
    \left\|
      u_2 + \tau v_2
    \right\|_{U_2}^2
    +
    \left\|
      u_2 - \tau v_2
    \right\|_{U_2}^2
    -
    2
    \left\|
      u_2
    \right\|_{U_2}^2
    -
    2 \tau^2
    \left\|
      v_2
    \right\|_{U_2}^2\\
    \le
    4
    \left\|
      u_2
    \right\|_{U_2}^2
    \rho_{U_2}
    \left(
      \tau
      \left\|
        v_2
      \right\|_{U_2}
      /
      \left\|
        u_2
      \right\|_{U_2}
    \right).
  \end{multline}
  Multiplying (\ref{eq:pseudo-parallelogram1}) by $a_1$
  and (\ref{eq:pseudo-parallelogram2}) by $a_2$
  and summing now gives
  \begin{multline}\label{eq:pseudo-next}
    \left\|
      u^{\dagger} + \tau v^{\dagger}
    \right\|_{U_1\oplus U_2}^2
    +
    \left\|
      u^{\dagger} - \tau v^{\dagger}
    \right\|_{U_1\oplus U_2}^2
    -
    2
    -
    2
    \tau^2\\
    \le
    4
    \sum_{j=1}^2
    a_j
    \left\|
      u_j
    \right\|_{U_j}^2
    \rho_{U_j}
    \left(
      \tau
      \left\|
        v_j
      \right\|_{U_j}
      /
      \left\|
        u_j
      \right\|_{U_j}
    \right).
  \end{multline}
  To estimate the sum over $j=1,2$,
  notice that:
  \begin{itemize}
  \item
    when $\left\|v_j\right\|_{U_j}\le\left\|u_j\right\|_{U_j}$,
    \begin{equation*}
      \rho_{U_j}
      \left(
        \tau
        \left\|
          v_j
        \right\|_{U_j}
        /
        \left\|
          u_j
        \right\|_{U_j}
      \right)
      \le
      \rho_{U_j}(\tau)
      \left\|
        v_j
      \right\|_{U_j}
      /
      \left\|
        u_j
      \right\|_{U_j}
    \end{equation*}
    (by the convexity of $\rho$,
    following from the convexity of the Fenchel transform,
    (\ref{eq:fenchel}),
    and the reflexivity of all uniformly convex and all uniformly smooth spaces);
  \item
    when $\left\|v_j\right\|_{U_j}>\left\|u_j\right\|_{U_j}$,
    \begin{equation*}
      \rho_{U_j}
      \left(
        \tau
        \left\|
          v_j
        \right\|_{U_j}
        /
        \left\|
          u_j
        \right\|_{U_j}
      \right)
      \le
      L
      \rho_{U_j}(\tau)
      \left(
        \left\|
          v_j
        \right\|_{U_j}
        /
        \left\|
          u_j
        \right\|_{U_j}
      \right)^2
    \end{equation*}
    (where $L<3.18$ is a constant satisfying $\rho(\sigma)/\sigma^2 \le L\rho(\tau)/\tau^2$
    for all positive $\tau\le\sigma$;
    see \cite{figiel:1976full}, Proposition 10 on p.~128 and the remark after its proof).
  \end{itemize}
  Using the Cauchy--Schwarz inequality,
  the sum can be bounded above as follows:
  \begin{multline}\label{eq:final-figiel}
    \sum_{j=1}^2
    a_j
    \left\|
      u_j
    \right\|_{U_j}^2
    \rho_{U_j}
    \left(
      \tau
      \left\|
        v_j
      \right\|_{U_j}
      /
      \left\|
        u_j
      \right\|_{U_j}
    \right)\\
    \le
    \sum_{j=1}^2
    a_j
    \left\|
      v_j
    \right\|_{U_j}
    \rho_{U_j}(\tau)
    \max
    \left(
      \left\|
        u_j
      \right\|_{U_j},
      L
      \left\|
        v_j
      \right\|_{U_j}
    \right)\\
    \le
    \left(
      \sum_{j=1}^2
      a_j
      \left\|
        v_j
      \right\|_{U_j}^2
    \right)^{1/2}
    \left(
      \sum_{j=1}^2
      a_j
      \left(
        \rho_{U_j}(\tau)
      \right)^2
      \left(
        \left\|
          u_j
        \right\|_{U_j}^2
        +
        L^2
        \left\|
          v_j
        \right\|_{U_j}^2
      \right)
    \right)^{1/2}\\
    \le
    \left(
      \sum_{j=1}^2
      f^2(\tau)
      a_j
      \left(
        \left\|
          u_j
        \right\|_{U_j}^2
        +
        L^2
        \left\|
          v_j
        \right\|_{U_j}^2
      \right)
    \right)^{1/2}
    =
    \sqrt{L^2+1}
    f(\tau)
  \end{multline}
  (the last line assuming $\tau\in(0,1]$).
  Now we have all we need to deduce the conclusion of the lemma
  (some steps will be explained after the equation):
  when $\tau\in(0,1]$,
  \begin{multline*}
    \frac12
    \left(
      \left\|
        u^{\dagger} + \tau v^{\dagger}
      \right\|_{U_1\oplus U_2}
      +
      \left\|
        u^{\dagger} - \tau v^{\dagger}
      \right\|_{U_1\oplus U_2}
    \right)\\
    \le
    \left(
      \frac12
      \left(
        \left\|
          u^{\dagger} + \tau v^{\dagger}
        \right\|_{U_1\oplus U_2}^2
        +
        \left\|
          u^{\dagger} - \tau v^{\dagger}
        \right\|_{U_1\oplus U_2}^2
      \right)
    \right)^{1/2}\\
    \le
    \left(
      1
      +
      \tau^2
      +
      2
      \sqrt{L^2+1}
      f(\tau)
    \right)^{1/2}
    \le
    \left(
      1
      +
      \tau^2
    \right)^{1/2}
    +
    \sqrt{L^2+1}
    f(\tau)\\
    \le
    1
    +
    f(\tau)
    +
    \sqrt{L^2+1}
    f(\tau)
    =
    1
    +
    \left(
      1
      +
      \sqrt{L^2+1}
    \right)
    f(\tau)
  \end{multline*}
  (the first inequality follows from the convexity of the function $t\mapsto t^2$,
  the second from (\ref{eq:pseudo-next}) and (\ref{eq:final-figiel}),
  the third from the mean-value theorem,
  and the fourth from N\"ordlander's bound (\ref{eq:Nordlander})).
  It remains to compare the resulting inequality
  with the definition of the modulus of convexity
  and remember that $L<3.18$.
  \qedtext
\end{proof}

\subsection*{Convexity and smoothness for Sobolev spaces}
\fi

It was shown by Clarkson \cite{clarkson:1936} (\S3) that, for $p\in[2,\infty)$,
\begin{equation*}
  \delta_{L^p}(\epsilon)
  \ge
  1
  -
  \left(
    1
    -
    (\epsilon/2)^p
  \right)^{1/p}.
\end{equation*}
(And this bound was shown to be optimal in \cite{hanner:1956}.)
A quick inspection
of the standard proofs
(see, e.g., \cite{\AdamsFournier}, 2.34--2.40)
shows that the underlying measurable space $\Omega$ and measure $\mu$ of $L^p=L^p(\Omega,\mu)$
can be essentially arbitrary
(only the degenerate case where $\dim L^p<2$ should be excluded),
although this generality is usually not emphasized.

It is easy to see
(cf.\ \cite{\AdamsFournier}, 3.5--3.6)
that the modulus of convexity of each Sobolev space $W^{s,p}(\mathbf{X})$,
$s\in(0,1)$ and $p\in[2,\infty)$,
also satisfies
\begin{equation}\label{eq:delta-Sobolev1}
  \delta_{W^{s,p}(\mathbf{X})}(\epsilon)
  \ge
  1
  -
  \left(
    1
    -
    (\epsilon/2)^p
  \right)^{1/p}.
\end{equation}
\ifnotCONF
Indeed, with each $f\in W^{s,p}(\mathbf{X})$
we can associate a function $\overline{f}:\mathbf{X}\cup\mathbf{X}^2\to\bbbr$
(we regard the sets $\mathbf{X}$ and $\mathbf{X}^2$ as disjoint)
such that
\begin{alignat*}{2}
  \overline{f}(x)
  &=
  f(x)&&
  \text{for }x\in\mathbf{X},
  \\
  \overline{f}(x,y)
  &=
  \frac{f(x)-f(y)}{\lvert x-y\rvert^s}&\quad&
  \text{for }(x,y)\in\mathbf{X}^2;
\end{alignat*}
the measure on $\mathbf{X}\cup\mathbf{X}^2$
coincides with the Lebesgue measure on the measurable subsets of $\mathbf{X}$
and with the measure whose density is
$(x,y)\in\mathbf{X}^2\mapsto\lvert x-y\rvert^{-m}$,
with respect to the Lebesgue measure,
on the measurable subsets of $\mathbf{X}^2$.
The bound (\ref{eq:delta-Sobolev1}) can now be deduced from Clarkson's result as follows:
\begin{multline*}
  \delta_{W^{s,p}(\mathbf{X})}(\epsilon)
  :=
  \inf_{\substack{f,g\in S_{W^{s,p}(\mathbf{X})}\\\left\|f-g\right\|_{W^{s,p}(\mathbf{X})}=\epsilon}}
  \left(
    1
    -
    \left\|
      \frac{f+g}{2}
    \right\|_{W^{s,p}(\mathbf{X})}
  \right)\\
  =
  \inf_{\substack{f,g:\mathbf{X}\to\bbbr\\
    \overline{f},\overline{g}\in L^{p}(\mathbf{X}\cup\mathbf{X}^2)\\
    \left\|\overline{f}-\overline{g}\right\|_{L^{p}(\mathbf{X}\cup\mathbf{X}^2)}=\epsilon}}
  \left(
    1
    -
    \left\|
      \frac{\overline{f}+\overline{g}}{2}
    \right\|_{L^{p}(\mathbf{X}\cup\mathbf{X}^2)}
  \right)\\
  \ge
  \inf_{\substack{u,v\in L^{p}(\mathbf{X}\cup\mathbf{X}^2)\\
    \left\|u-v\right\|_{L^{p}(\mathbf{X}\cup\mathbf{X}^2)}=\epsilon}}
  \left(
    1
    -
    \left\|
      \frac{u+v}{2}
    \right\|_{L^{p}(\mathbf{X}\cup\mathbf{X}^2)}
  \right)\\
  =
  \delta_{L^{p}(\mathbf{X}\cup\mathbf{X}^2)}(\epsilon)
  \ge
  1
  -
  \left(
    1
    -
    (\epsilon/2)^p
  \right)^{1/p}.
\end{multline*}

\fi
Since, for $t\in[0,1]$ and $p\ge1$,
$
  (1-t)^{1/p}
  \le
  1 - t/p
$
(the left-hand side is a concave function of $t$,
and the values and derivatives of the two sides match when $t=0$),
we have
\begin{equation}\label{eq:delta-Sobolev2}
  \delta_{W^{s,p}(\mathbf{X})}(\epsilon)
  \ge
  (\epsilon/2)^p/p.
\end{equation}
Therefore, as we said in \S\ref{sec:main},
the Sobolev spaces indeed satisfy the condition (\ref{eq:condition}) of Theorem \ref{thm:main}.

\section{Proof of Theorem \ref{thm:main}}
\label{sec:proof}

In this section we partly follow the proof of Theorem~1 in \cite{\GTPXI} (\S6).

\subsection*{The BBK29 algorithm}

Let $U$ be a Banach space.
We say that a function $\Phi:[-Y,Y]\times\mathbf{X}\to U$ is \emph{forecast-continuous}
if $\Phi(\mu,x)$ is continuous in $\mu\in[-Y,Y]$ for every fixed $x\in\mathbf{X}$.
For such a $\Phi$ the function
\begin{multline}\label{eq:function}
  f_n(y,\mu)
  :=
  \left\|
    \sum_{i=1}^{n-1}
    (y_i-\mu_i)
    \Phi
    \bigl(
      \mu_i,x_i
    \bigr)
    +
    (y-\mu)
    \Phi
    \bigl(
      \mu,x_n
    \bigr)
  \right\|_U\\
  -
  \left\|
    \sum_{i=1}^{n-1}
    (y_i-\mu_i)
    \Phi
    \bigl(
      \mu_i,x_i
    \bigr)
  \right\|_U
\end{multline}
is continuous in $\mu\in[-Y,Y]$.

\bigskip

\noindent
\textsc{Banach-space Balanced K29 algorithm (BBK29)}

\noindent
\textbf{Parameter:} forecast-continuous $\Phi:[-Y,Y]\times\mathbf{X}\to U$,
  with $U$ a Banach space

\parshape=8
\IndentI   \WidthI
\IndentII  \WidthII
\IndentII  \WidthII
\IndentII  \WidthII
\IndentIII \WidthIII
\IndentIII \WidthIII
\IndentII  \WidthII
\IndentI   \WidthI
\noindent
FOR $n=1,2,\dots$:\\
  Read $x_n\in\mathbf{X}$.\\
  Define $f_n:[-Y,Y]^2\to\bbbr$ by (\ref{eq:function}).\\
  Output any root $\mu\in[-Y,Y]$ of $f_n(-Y,\mu)=f_n(Y,\mu)$ as $\mu_n$;\\
    if there are no such roots, output $\mu_n\in\{-Y,Y\}$\\
    such that $\sup_{y\in[-Y,Y]}f_n(y,\mu_n)\le0$.\\
  Read $y_n\in[-Y,Y]$.\\
END FOR.

\bigskip

\noindent
The validity of this description
depends on the existence of $\mu\in\{-Y,Y\}$ satisfying $\sup_{y\in[-Y,Y]}f_n(y,\mu)\le0$
when the equation $f_n(-Y,\mu)=f_n(Y,\mu)$ does not have roots $\mu\in[-Y,Y]$.
The existence of such a $\mu$ is easy to check:
if $f_n(-Y,\mu)<f_n(Y,\mu)$ for all $\mu\in[-Y,Y]$,
take $\mu:=Y$ to obtain
\begin{equation*}
  f_n(-Y,\mu)<f_n(Y,\mu)=0
\end{equation*}
and, hence, $\sup_{y\in[-Y,Y]}f_n(y,\mu)\le0$ by the convexity of (\ref{eq:function}) in $y$;
if $f_n(-Y,\mu)>f_n(Y,\mu)$ for all $\mu\in[-Y,Y]$,
setting $\mu:=-Y$ leads to
\begin{equation*}
  f_n(Y,\mu)<f_n(-Y,\mu)=0
\end{equation*}
and, hence, $\sup_{y\in[-Y,Y]}f_n(y,\mu)\le0$.
The parameter $\Phi$ of the BBK29 algorithm
will sometimes be called the \emph{feature mapping}.
\begin{theorem}\label{thm:K29}
  Let $\Phi$ be a forecast-continuous mapping
  from $[-Y,Y]\times\mathbf{X}$
  to a Banach space $U$
  and set
  $\ccc_{\Phi}:=\sup_{\mu\in[-Y,Y],x\in\mathbf{X}}\left\|\Phi(\mu,x)\right\|_U$.
  Suppose $\rho_U(\tau)\le a\tau^q$, $\forall\tau\in(0,1]$,
  for some constants $q\ge1$ and $a\ge1/q$.
  The BBK29 algorithm with parameter $\Phi$ outputs $\mu_n\in[-Y,Y]$
  such that
  \begin{equation}\label{eq:K29}
    \left\|
      \sum_{n=1}^N
      (y_n-\mu_n)
      \Phi(\mu_n,x_n)
    \right\|_{U}
    \le
    2Y\ccc_{\Phi}
    \left(
      2aq
      N
    \right)^{1/q}
  \end{equation}
  always holds for all $N=1,2,\dots$.
\end{theorem}
\ifnotCONF
\begin{proof}
  Set
  \begin{equation*}
    S_N
    :=
    \left\|
      \sum_{n=1}^N
      (y_n-\mu_n)
      \Phi(\mu_n,x_n)
    \right\|_{U};
  \end{equation*}
  our goal is to prove
  \begin{equation*}
    S_N
    \le
    2Y\ccc_{\Phi}
    \left(
      2aq
      N
    \right)^{1/q}.
  \end{equation*}
  For $N=1$,
  this follows from
  \begin{equation*}
    2Y\ccc_{\Phi}
    \le
    2Y\ccc_{\Phi}
    \left(
      2aq
      N
    \right)^{1/q},
  \end{equation*}
  which in turn follows from $2aq\ge1$,
  which in turn follows from the condition $a\ge1/q$.
  It remains to prove that
  \begin{equation*}
    S_{N-1}
    \le
    2Y\ccc_{\Phi}
    \left(
      {2a}{q}
      (N-1)
    \right)^{1/q}
  \end{equation*}
  implies
  \begin{equation}\label{eq:S-N-above}
    S_N
    \le
    2Y\ccc_{\Phi}
    \left(
      {2a}{q}
      N
    \right)^{1/q}
  \end{equation}
  for $N\ge2$.
  Without loss of generality we assume that $f_N(-Y,\mu_N)=f_N(Y,\mu_N)$
  and replace $S_N$ in (\ref{eq:S-N-above}) by $f_N:=f_N(Y,\mu_N)$.

  Fix $N\ge2$.
  We will assume that
  \begin{equation}\label{eq:contradictory}
    S_{N-1}
    \le
    2Y\ccc_{\Phi}
    \left(
      {2a}{q}
      (N-1)
    \right)^{1/q}
    \quad\&\quad
    f_N
    >
    2Y\ccc_{\Phi}
    \left(
      {2a}{q}
      N
    \right)^{1/q}
  \end{equation}
  and arrive at a contradiction.
  By the definition of $\rho^{\ddagger}$,
  \begin{equation*}
    S_{N-1}
    \ge
    f_N
    \left(
      1
      -
      \rho^{\ddagger}_U
      \left(
        \frac{2Y\left\|\Phi(\mu_N,x_N)\right\|}{f_N}
      \right)
    \right)
  \end{equation*}
  (cf.\ Figure \ref{fig:smoothness}).
  Since $f_N>2Y\ccc_{\Phi}$
  (remember that we are assuming (\ref{eq:contradictory})),
  by Corollary~\ref{cor:rho-0-1} and Lemma~\ref{lem:rho-1-2}
  this implies
  \begin{equation*}
    S_{N-1}
    \ge
    f_N
    \left(
      1
      -
      2a
      \left(
        \frac{2Y\left\|\Phi(\mu_N,x_N)\right\|}{f_N}
      \right)^q
    \right).
  \end{equation*}
  As the right-hand side is a monotonically increasing function of $f_N$
  (which can be checked by differentiation),
  in combination with (\ref{eq:contradictory}) the last inequality gives
  \begin{equation*}
    2Y\ccc_{\Phi}
    \left(
      {2a}{q}
      (N-1)
    \right)^{1/q}
    >
    2Y\ccc_{\Phi}
    \left(
      {2a}{q}
      N
    \right)^{1/q}
    \left(
      1
      -
      2a
      \left(
        \left(
          {2a}{q}
          N
        \right)^{-1/q}
      \right)^q
    \right),
  \end{equation*}
  i.e.,
  \begin{equation*}
    (N-1)^{1/q}
    >
    N^{1/q}
    \left(
      1
      -
      \frac{1}{qN}
    \right).
  \end{equation*}
  It remains to rewrite the last inequality as
  \begin{equation}\label{eq:last-inequality}
    N^{1/q}
    -
    (N-1)^{1/q}
    <
    \frac{1}{q}
    N^{1/q-1}
  \end{equation}
  and notice that, by the mean-value theorem,
  the left-hand side of (\ref{eq:last-inequality}) equals
  \begin{equation*}
    \frac{1}{q}
    (N-\theta)^{1/q-1}
  \end{equation*}
  for some $\theta\in(0,1)$:
  as $1/q-1\le0$,
  we have the required contradiction.
  \qedtext
\end{proof}
\fi

\subsection*{The feature mapping for the proof of Theorem \ref{thm:main}}

In the proof of Theorem \ref{thm:main}
we will need two feature mappings from $[-Y,Y]\times\mathbf{X}$ to different Banach spaces:
first, $\Phi_1(\mu,x):=\mu$
(mapping to the Banach space $\bbbr$),
and second, $\Phi_2:[-Y,Y]\times\mathbf{X}\to\FFF^*$
such that $\Phi_2(\mu,x)$ is the evaluation functional $\kkk_x:f\mapsto f(x)$,
$f\in\FFF$.
We combine them into one feature mapping
\begin{equation}\label{eq:combination}
  \Phi(\mu,x)
  :=
  \bigl(
    \Phi_1(\mu,x),
    \Phi_2(\mu,x)
  \bigr)
\end{equation}
to the weighted direct sum $U:=\bbbr\oplus\FFF^*$,
with the weights $a_1$ and $a_2$ to be chosen later.
By Lemma \ref{lem:direct}, (\ref{eq:dual}), and (\ref{eq:Hilbert}),
$\rho_U(\tau)\le a\tau^q$,
where $a:=4.34/q$.
With the help of Theorem \ref{thm:K29},
we obtain for the BBK29 algorithm with parameter $\Phi$:
\begin{multline}\label{eq:calibration}
  \left|
    \sum_{n=1}^N
    (y_n-\mu_n)
    \mu_n
  \right|
  =
  \left\|
    \sum_{n=1}^N
    (y_n-\mu_n)
    \Phi_1(\mu_n,x_n)
  \right\|_{\bbbr}\\
  \le
  \frac{1}{\sqrt{a_1}}
  \left\|
    \sum_{n=1}^N
    (y_n-\mu_n)
    \Phi(\mu_n,x_n)
  \right\|_{U}
  \le
  \frac{1}{\sqrt{a_1}}
  2Y\ccc_{\Phi}
  \left(
    2aqN
  \right)^{1/q}
\end{multline}
and
\begin{multline}\label{eq:resolution}
  \left|
    \sum_{n=1}^N
    (y_n-\mu_n)
    D(x_n)
  \right|
  =
  \left|
    \sum_{n=1}^N
    (y_n-\mu_n)
    \kkk_{x_n}(D)
  \right|
  =
  \left|
    \left(
      \sum_{n=1}^N
      (y_n-\mu_n)
      \kkk_{x_n}
    \right)
    (D)
  \right|\\
  \le
  \left\|
    \sum_{n=1}^N
    (y_n-\mu_n)
    \kkk_{x_n}
  \right\|_{\FFF^*}
  \left\|
    D
  \right\|_{\FFF}
  =
  \left\|
    \sum_{n=1}^N
    (y_n-\mu_n)
    \Phi_2(\mu_n,x_n)
  \right\|_{\FFF^*}
  \left\|
    D
  \right\|_{\FFF}\\
  \le
  \frac{1}{\sqrt{a_2}}
  \left\|
    \sum_{n=1}^N
    (y_n-\mu_n)
    \Phi(\mu_n,x_n)
  \right\|_{U}
  \left\|
    D
  \right\|_{\FFF}
  \le
  \frac{1}{\sqrt{a_2}}
  2Y\ccc_{\Phi}
  \left(
    2aqN
  \right)^{1/q}
  \left\|
    D
  \right\|_{\FFF}
\end{multline}
for each function $D\in\FFF$.

\subsection*{Proof proper}

The proof is based on the inequality
\begin{align*}
  &\sum_{n=1}^N
  (y_n-\mu_n)^2\notag\\
  &=
  \sum_{n=1}^N
  (y_n-D(x_n))^2
  +
  2
  \sum_{n=1}^N
  (D(x_n)-\mu_n)
  (y_n-\mu_n)
  -
  \sum_{n=1}^N
  (D(x_n)-\mu_n)^2\notag\\
  &\le
  \sum_{n=1}^N
  (y_n-D(x_n))^2
  +
  2
  \sum_{n=1}^N
  (D(x_n)-\mu_n)
  (y_n-\mu_n)\notag
\end{align*}
(immediately following from (\ref{eq:elementary})).
Using this inequality
and (\ref{eq:calibration})--(\ref{eq:resolution})
with $a_1:=Y^{-2}$ and $a_2:=1$,
we obtain for the $\mu_n\in[-Y,Y]$
output by the BBK29 algorithm with $\Phi$ as parameter:
\begin{align*}
  &\sum_{n=1}^N
  (y_n-\mu_n)^2\\
  &\le
  \sum_{n=1}^N
  (y_n-D(x_n))^2
  +
  2
  \left|
    \sum_{n=1}^N
    \mu_n
    (y_n-\mu_n)
  \right|
  +
  2
  \left|
    \sum_{n=1}^N
    D(x_n)
    (y_n-\mu_n)
  \right|\\
  &\le
  \sum_{n=1}^N
  (y_n-D(x_n))^2
  +
  4Y\ccc_{\Phi}
  \left(
    2aqN
  \right)^{1/q}
  \left(
    \left\|D\right\|_{\FFF} + Y
  \right).
\end{align*}
Since
\begin{equation*}
  \ccc_{\Phi}
  \le
  \sqrt{a_1Y^2+a_2\ccc_{\FFF}^2}
  =
  \sqrt{\ccc_{\FFF}^2+1},
\end{equation*}
we can see that (\ref{eq:main}) holds with
\begin{equation}\label{eq:C}
  4 (2aq)^{1/q}
  =
  4
  \times
  8.68^{1/q}
\end{equation}
in place of $40$.

\section{Banach kernels}

An RKHS can be defined as a PBFS in which the norm
is expressed via an inner product as $\left\|f\right\|=\surd\left\langle f,f\right\rangle$.
It is well known that all information about an RKHS $\FFF$ on $Z$
is contained in its ``reproducing kernel'',
which is a symmetric positive definite function on $Z^2$
(\cite{aronszajn:1950}, \S\S I.1--I.2).
The reproducing kernel can be regarded as the constructive representation
of its RKHS,
and it is the reproducing kernel rather than the RKHS itself
that serves as a parameter of various machine-learning algorithms.
In this section we will introduce a similar constructive representation
for PBFS.

A \emph{Banach kernel} $B$ on a set $Z$
is a function that maps each finite non-empty sequence $z_1,\ldots,z_n$
of distinct elements of $Z$
to a seminorm $\left\|\cdot\right\|_{B(z_1,\ldots,z_n)}$ on $\bbbr^n$
and satisfies the following conditions
(familiar from Kolmogorov's existence theorem
\cite{kolmogorov:1933}, \S III.4):
\begin{itemize}
\item
  for each $n=1,2,\ldots$,
  each sequence $z_1,\ldots,z_n$ of distinct elements of $Z$,
  each sequence $(t_1,\ldots,t_n)\in\bbbr^n$,
  and each permutation
  $
   \bigl(
     \begin{smallmatrix}
       1&2&\ldots&n\\
       i_1&i_2&\ldots&i_n
     \end{smallmatrix}
   \bigr)
  $,
  \begin{equation*}
    \left\|
      \left(
        t_{i_1},\ldots,t_{i_n}
      \right)
    \right\|_{B(z_{i_1},\ldots,z_{i_n})}
    =
    \left\|
      \left(
        t_1,\ldots,t_n
      \right)
    \right\|_{B(z_1,\ldots,z_n)};
  \end{equation*}
\item
  for each $n=1,2,\ldots$, each $k=1,\ldots,n$,
  each sequence $z_1,\ldots,z_n$ of distinct elements of $Z$,
  and each sequence $(t_1,\ldots,t_k)\in\bbbr^k$,
  \begin{equation*}
    \left\|
      \left(
        t_1,\ldots,t_k
      \right)
    \right\|_{B(z_1,\ldots,z_k)}
    =
    \left\|
      \left(
        t_1,\ldots,t_k,0,\ldots,0
      \right)
    \right\|_{B(z_1,\ldots,z_n)}.
  \end{equation*}
\end{itemize}
\ifnotCONF

The \emph{Banach kernel of a mapping} $\Phi:Z\to U$
to a Banach space $U$
is the Banach kernel $B$ defined by
\begin{equation*}
  \left\|
    \left(
      t_1,\ldots,t_n
    \right)
  \right\|_{B(z_1,\ldots,z_n)}
  :=
  \left\|
    t_1\Phi(z_1)+\cdots+t_n\Phi(z_n)
  \right\|_{U}.
\end{equation*}

\begin{proposition}\label{prop:kernel1}
  For each Banach kernel $B$ on $Z$
  there exists a Banach space $U$
  and a mapping $\Phi:Z\to U$
  such that $B$ is the Banach kernel of $\Phi$.
\end{proposition}
Proposition \ref{prop:kernel1} is a special case
of the following Proposition \ref{prop:kernel2},
but we still need to prove it
as the proof of Proposition \ref{prop:kernel2} depends on it.
\begin{Proof}{of Proposition \ref{prop:kernel1}}
  Let $U_1$ be the set of all formal linear combinations
  $t_1z_1+\cdots+t_nz_n$,
  where $n\in\{0,1,2,\ldots\}$,
  $(t_1,\ldots,t_n)\in(\bbbr\setminus\{0\})^n$,
  and $z_1,\ldots,z_n$ are distinct elements of $Z$.
  (There is only one linear combination,
  denoted $0$,
  corresponding to $n=0$.)
  We do not distinguish linear combinations
  if they have the same addends
  (perhaps listed in different orders).
  The set $U_1$ is a linear space
  with the obvious operations of addition
  and multiplication by scalar:
  in the sum the addends that are multiples of the same $z\in Z$
  should be grouped together
  (and removed if the resulting coefficient is zero)
  and multiplication by $0$ gives $0$.

  For each linear combination $t_1z_1+\cdots+t_nz_n\in U_1$, $n>0$,
  its seminorm is defined to be
  $
    \left\|
      \left(
        t_1,\ldots,t_n
      \right)
    \right\|_{B(z_1,\ldots,z_n)}
  $,
  and the seminorm of $0\in U_1$ is defined to be $0$;
  it is easy to check that this is indeed a seminorm
  (it is well defined because of the first condition
  in the definition of Banach kernel,
  and the triangle inequality follows from the second condition).
  Two linear combinations
  are said to be \emph{equivalent}
  if their difference has zero seminorm
  (this is indeed an equivalence relation because of the second condition).
  Let $U_2$ be the set of all equivalence classes.

  The norm of $u\in U_2$ can be defined
  as the seminorm of any element of the equivalence class $u$.
  It remains to take the completion of $U_2$ as $U$
  and to define $\Phi:Z\to U$ so that $\Phi(z)$ is the equivalence class
  containing $1z\in U_1$.
  \qedtext
\end{Proof}

\fi
The \emph{Banach kernel of a PBFS} $\FFF$ on $Z$
is the Banach kernel $B$ defined by
\begin{equation*}
  \left\|
    \left(
      t_1,\ldots,t_n
    \right)
  \right\|_{B(z_1,\ldots,z_n)}
  :=
  \left\|
    t_1\kkk_{z_1}+\cdots+t_n\kkk_{z_n}
  \right\|_{\FFF^*},
\end{equation*}
where $\kkk_z:\FFF\to\bbbr$, $z \in Z$,
is the evaluation functional $f\in\FFF\mapsto f(z)$.

\begin{proposition}\label{prop:kernel2}
  For each Banach kernel $B$ on $Z$
  there exists a proper Banach functional space $\FFF$ on $Z$
  such that $B$ is the Banach kernel of $\FFF$.
\end{proposition}

\ifnotCONF
\begin{proof}
  Let $\Phi:Z\to U$ be a mapping to a Banach space $U$
  such that $B$ is the Banach kernel of $\Phi$
  (such a $\Phi$ exists by Proposition \ref{prop:kernel1}).
  Without loss of generality we will assume
  that $\Phi(Z)$ spans $U$.
  Define $\FFF$ to be the set of all functions $f:Z\to\bbbr$
  of the form
  \begin{equation}\label{eq:form}
    f(z)
    :=
    \phi(\Phi(z)),
  \end{equation}
  where $\phi$ is a continuous linear functional on $U$,
  $\phi\in U^*$.
  The norm of the function (\ref{eq:form}) is
  $
    \left\|
      f
    \right\|_{\FFF}
    :=
    \left\|
      \phi
    \right\|_{U^*}
  $.
  We will prove that $\FFF$ is a PBFS
  and that $B$ is the Banach kernel of $\FFF$.

  It is obvious that $\FFF$ is a linear space
  (under the usual pointwise operations
  of addition and multiplication by scalar)
  and that $\left\|f\right\|_{\FFF}$ is well-defined
  (i.e., does not depend on the choice of $\phi$
  satisfying (\ref{eq:form}):
  there is only one such $\phi$).
  All defining properties of a norm are clearly satisfied
  for $\left\|\cdot\right\|_{\FFF}$;
  in particular, $\left\|f\right\|_{\FFF}=0$ implies $f=0$.
  The completeness of $\FFF$ follows from the completeness of $U^*$.
  The boundedness of the evaluation functionals for $\FFF$
  means that, for each fixed $z\in Z$,
  \begin{equation*}
    \sup_{\phi:\left\|\phi\right\|_{U^*}\le1}
    \left|
      \phi(\Phi(z))
    \right|
    <
    \infty;
  \end{equation*}
  this immediately follows from the definition of $\left\|\cdot\right\|_{U^*}$.
  This completes the proof that $\FFF$ is a PBFS.

  It remains to check that $B$ is the Banach kernel of $\FFF$,
  i.e., that
  \begin{equation}\label{eq:to-prove}
    \left\|
      \left(
        t_1,\ldots,t_n
      \right)
    \right\|_{B(z_1,\ldots,z_n)}
    =
    \left\|
      \phi
      \mapsto
      t_1\phi(\Phi(z_1))+\cdots+t_n\phi(\Phi(z_n))
    \right\|_{U^{**}}
  \end{equation}
  for all $n=1,2,\ldots$, all $(t_1,\ldots,t_n)\in(\bbbr\setminus\{0\})^n$,
  and all distinct $z_1,\ldots,z_n\in Z$.
  We can rewrite (\ref{eq:to-prove}) as
  \begin{equation*}
    \left\|
      \left(
        t_1,\ldots,t_n
      \right)
    \right\|_{B(z_1,\ldots,z_n)}
    =
    \left\|
      \phi
      \mapsto
      \phi
      \left(
        t_1\Phi(z_1)+\cdots+t_n\Phi(z_n)
      \right)
    \right\|_{U^{**}};
  \end{equation*}
  since $B$ is the Banach kernel of $\Phi$,
  this is equivalent to
  \begin{equation*}
    \left\|
      t_1\Phi(z_1)+\cdots+t_n\Phi(z_n)
    \right\|_{U}
    =
    \left\|
      \phi
      \mapsto
      \phi
      \left(
        t_1\Phi(z_1)+\cdots+t_n\Phi(z_n)
      \right)
    \right\|_{U^{**}}.
  \end{equation*}
  The last equality follows from the fact that the canonical imbedding
  of $U$ into $U^{**}$ is an isometry
  (\cite{rudin:1991}, \S4.5).
  \qedtext
\end{proof}

\begin{remark*}
  A Banach kernel $B$ on $Z$ can be visualized as a family
  $b(z_1,\ldots,z_n)\subseteq\bbbr^n$,
  $n$ ranging over $\{1,2,\ldots\}$
  and $z_1,\ldots,z_n$ over sequences of distinct elements of $Z$,
  of balanced convex sets containing a neighborhood of zero.
  Such a family can be obtained from $B$ by replacing each seminorm
  $\left\|\cdot\right\|_{B(z_1,\ldots,z_n)}$
  with the unit ball in that seminorm;
  it is well known that the seminorm and the corresponding unit ball
  carry the same information
  (see, e.g., \cite{rudin:1991}, Theorems 1.34 and 1.35).
  Of course,
  the sets $b(z_1,\ldots,z_n)$ should satisfy the two conditions of consistency
  analogous to those in the definition of a Banach kernel;
  e.g., the second condition becomes:
  for all $n=1,2,\ldots$, all $k=1,\ldots,n$,
  and all $(z_1,\ldots,z_n)\in Z^n$ whose elements are all different,
  the set $b(z_1,\ldots,z_k)$ is the intersection of $b(z_1,\ldots,z_n)$
  and the hyperplane $z_{k+1}=\cdots=z_n=0$.
\end{remark*}
\fi

Now we can state more explicitly the prediction algorithm
described above and guaranteeing (\ref{eq:main}).
Following (\ref{eq:function}) (with $\Phi$ defined by (\ref{eq:combination})),
define
\begin{multline}\label{eq:kernel-function}
  f_n(y,\mu)
  :=
  \Biggl(
    \frac{1}{Y^2}
    \left(
      \sum_{i=1}^{n-1}
      (y_i-\mu_i)
      \mu_i
      +
      (y-\mu)
      \mu
    \right)^2\\
    +
    \left\|
      \left(
        y_1-\mu_1,
        \ldots,
        y_{n-1}-\mu_{n-1},
        y-\mu
      \right)
    \right\|_{B(x_1,\ldots,x_{n-1},x_n)}^2
  \Biggr)^{1/2}\\
  -
  \Biggl(
    \frac{1}{Y^2}
    \left(
      \sum_{i=1}^{n-1}
      (y_i-\mu_i)
      \mu_i
    \right)^2\\
    +
    \left\|
      \left(
        y_1-\mu_1,
        \ldots,
        y_{n-1}-\mu_{n-1}
      \right)
    \right\|_{B(x_1,\ldots,x_{n-1})}^2
  \Biggr)^{1/2}.
\end{multline}
This allows us to give the kernel representation of BBK29
with $\Phi$ defined by (\ref{eq:combination});
its parameter is a Banach kernel on the object space $\mathbf{X}$.

\bigskip

\noindent
\textsc{Algorithm guaranteeing (\ref{eq:main})}

\noindent
\textbf{Parameter:} Banach kernel $B$ of $\FFF$

\parshape=8
\IndentI   \WidthI
\IndentII  \WidthII
\IndentII  \WidthII
\IndentII  \WidthII
\IndentIII \WidthIII
\IndentIII \WidthIII
\IndentII  \WidthII
\IndentI   \WidthI
\noindent
FOR $n=1,2,\dots$:\\
  Read $x_n\in\mathbf{X}$.\\
  Define $f_n:[-Y,Y]^2\to\bbbr$ by (\ref{eq:kernel-function}).\\
  Output any root $\mu\in[-Y,Y]$ of $f_n(-Y,\mu)=f_n(Y,\mu)$ as $\mu_n$;\\
    if there are no such roots, output $\mu_n\in\{-Y,Y\}$\\
    such that $\sup_{y\in[-Y,Y]}f_n(y,\mu_n)\le0$.\\
  Read $y_n\in[-Y,Y]$.\\
END FOR.

\subsection*{Acknowledgments}

I am grateful to Glenn Shafer for a series of useful discussions.
This work was partially supported
by MRC (grant S505/65) and the Royal Society.


\begin{thebibliography}{10}

\bibitem{adams:1975}
Robert~A\DOT{} Adams.
\newblock {\em Sobolev Spaces}, volume~65 of {\em Pure and Applied
  Mathematics}.
\newblock Academic Press, New York, first edition, 1975.

\bibitem{adams/fournier:2003full}
Robert~A\DOT{} Adams and John J\DOT{}~F\DOT{} Fournier.
\newblock {\em Sobolev Spaces}, volume 140 of {\em Pure and Applied
  Mathematics}.
\newblock Academic Press, Amsterdam, second edition, 2003.
\newblock This new edition is not a superset of \cite{adams:1975}: some less
  important material is deleted.

\bibitem{aronszajn:1950}
Nachman Aronszajn.
\newblock Theory of reproducing kernels.
\newblock {\em Transactions of the American Mathematical Society}, 68:337--404,
  1950.

\bibitem{auer/etal:2002}
Peter Auer, Nicol\`o Cesa-Bianchi, and Claudio Gentile.
\newblock Adaptive and self-confident on-line learning algorithms.
\newblock {\em Journal of Computer and System Sciences}, 64:48--75, 2002.

\bibitem{banas:1986}
J\'ozef Bana\'s.
\newblock On moduli of smoothness of {B}anach spaces.
\newblock {\em Bulletin of the Polish Academy of Sciences. Mathematics},
  34:287--293, 1986.

\bibitem{banas/fraczek:1993}
J\'ozef Bana\'s and Krzysztof Fr\c{a}czek.
\newblock Deformation of {B}anach spaces.
\newblock {\em Commentationes Mathematicae Universitatis Carolinae}, 34:47--53,
  1993.

\bibitem{barcenas/etal:2004}
Di\'omedes B\'arcenas, Vladimir~I\DOT{} Gurary, Luisa S\'anchez, and Antonio
  Ull\'an.
\newblock On moduli of convexity in {B}anach spaces.
\newblock {\em Quaestiones Mathematicae}, 27:137--145, 2004.

\bibitem{cesabianchi/etal:2004}
Nicol\`o Cesa-Bianchi, Alex Conconi, and Claudio Gentile.
\newblock On the generalization ability of on-line learning algorithms.
\newblock {\em IEEE Transactions on Information Theory}, 50:2050--2057, 2004.

\bibitem{cesabianchi/long/warmuth:1996}
Nicol\`o Cesa-Bianchi, Philip~M\DOT{} Long, and Manfred~K\DOT{} Warmuth.
\newblock Worst-case quadratic loss bounds for on-line prediction of linear
  functions by gradient descent.
\newblock {\em IEEE Transactions on Neural Networks}, 7:604--619, 1996.

\bibitem{clarkson:1936}
James~A\DOT{} Clarkson.
\newblock Uniformly convex spaces.
\newblock {\em Transactions of the American Mathematical Society}, 40:396--414,
  1936.

\bibitem{devroye/etal:1996}
Luc Devroye, L\'aszl\'o Gy\"orfi, and G\'abor Lugosi.
\newblock {\em A Probabilistic Theory of Pattern Recognition}, volume~31 of
  {\em Applications of Mathematics}.
\newblock Springer, New York, 1996.

\bibitem{figiel:1976full}
T\DOT{} Figiel.
\newblock On the moduli of convexity and smoothness.
\newblock {\em Studia Mathematica}, 56:121--155, 1976.
\newblock Available free of charge at \texttt{http://matwbn.icm.edu.pl}.

\bibitem{fuster:2005}
E\DOT{}~Llorens Fuster.
\newblock Moduli and constants: \ldots what a show!
\newblock Available on the Internet (accessed in November 2005), May 2005.

\bibitem{gurary:1967latin}
Vladimir~I\DOT{} Gurary.
\newblock On differential properties of the complexity moduli of {B}anach
  spaces (in {R}ussian).
\newblock {\em Matematicheskie Issledovaniya}, 2:141--148, 1967.

\bibitem{hanner:1956}
Olof Hanner.
\newblock On the uniform convexity of {$L^p$} and {$l^p$}.
\newblock {\em Arkiv f\"or Matematik}, 3:239--244, 1956.

\bibitem{hoeffding:1963}
Wassily Hoeffding.
\newblock Probability inequalities for sums of bounded random variables.
\newblock {\em Journal of the American Statistical Association}, 58:13--30,
  1963.

\bibitem{kalman:1960}
Rudolph~E\DOT{} Kalman.
\newblock A new approach to linear filtering and prediction problems.
\newblock {\em Transactions of the ASME---Journal of Basic Engineering},
  82D:35--45, 1960.

\bibitem{kalman/bucy:1961}
Rudolph~E\DOT{} Kalman and Richard~S\DOT{} Bucy.
\newblock New results in linear filtering and prediction theory.
\newblock {\em Transactions of the ASME---Journal of Basic Engineering},
  83D:95--108, 1961.

\bibitem{karatzas/shreve:1991}
Ioannis Karatzas and Steven~E\DOT{} Shreve.
\newblock {\em Brownian Motion and Stochastic Calculus}.
\newblock Springer, New York, second edition, 1991.

\bibitem{kivinen/warmuth:1997}
Jyrki Kivinen and Manfred~K\DOT{} Warmuth.
\newblock {E}xponential {G}radient versus {G}radient {D}escent for linear
  predictors.
\newblock {\em Information and Computation}, 132:1--63, 1997.

\bibitem{kolmogorov:1933}
Andrei~N\DOT{} Kolmogorov.
\newblock {\em Grundbegriffe der Wahr\-schein\-lich\-keits\-rechnung}.
\newblock Springer, Berlin, 1933.
\newblock English translation (1950): \emph{Foundations of the theory of
  probability}. Chelsea, New York.

\bibitem{lindenstrauss:1963}
Joram Lindenstrauss.
\newblock On the modulus of smoothness and divergent series in {B}anach spaces.
\newblock {\em Michigan Mathematical Journal}, 10:241--252, 1963.

\bibitem{lindenstrauss/tzafriri:1979}
Joram Lindenstrauss and Lior Tzafriri.
\newblock {\em Classical Banach Spaces II: Function Spaces}, volume~97 of {\em
  Ergebnisse der Mathematik und ihrer Grenzgebiete}.
\newblock Springer, Berlin, 1979.

\bibitem{liokumovich:1973latin}
V\DOT{}~I\DOT{} Liokumovich.
\newblock The existence of {$B$}-spaces with non-convex modulus of convexity
  (in {R}ussian).
\newblock {\em Izvestiya Vysshikh Uchebnykh Zavedenii. Matematika}, 12:43--50,
  1973.

\bibitem{liptser/shiryaev:1974latin}
Robert~S\DOT{} Liptser and Albert~N\DOT{} Shiryaev.
\newblock {\em Sta\-tis\-ti\-ka\ slu\-chai\-nykh pro\-tses\-sov}.
\newblock Nauka, Moscow, 1974.
\newblock English translation: \emph{Statistics of Random Processes}. Springer,
  New York. In two volumes: \emph{General Theory} (1977) and
  \emph{Applications} (1978).

\bibitem{marti:1983}
J\DOT{}~T\DOT{} Marti.
\newblock Evaluation of the least constant in {S}obolev's inequality for
  {$H^1(0,s)$}.
\newblock {\em SIAM Journal on Numerical Analysis}, 20:1239--1242, 1983.

\bibitem{nikolsky:1961latin}
Sergei~M\DOT{} Nikolsky.
\newblock On imbedding, continuation and approximation theorems for
  differentiable functions of several variables.
\newblock {\em Russian Mathematical Surveys}, 16(5):55--104, 1961.
\newblock Russian original in: \emph{Uspekhi matemati\-ches\-kikh nauk},
  16(5):63--114.

\bibitem{nordlander:1960}
G\DOT{} N\"ordlander.
\newblock The modulus of convexity in normed linear spaces.
\newblock {\em Arkiv f\"or Matematik}, 4:15--17, 1960.

\bibitem{nuzman/poor:2000}
Carl~J\DOT{} Nuzman and H\DOT{}~Vincent Poor.
\newblock Linear estimation of self-similar processes via {L}amperti's
  transformation.
\newblock {\em Journal of Applied Probability}, 37:429--452, 2000.

\bibitem{rockafellar:1970}
R\DOT{}~Tyrrell Rockafellar.
\newblock {\em Convex Analysis}.
\newblock Princeton University Press, Princeton, NJ, 1970.

\bibitem{rudin:1991}
Walter Rudin.
\newblock {\em Functional Analysis}.
\newblock McGraw-Hill, Boston, second edition, 1991.

\bibitem{shafer/vovk:2001}
Glenn Shafer and \Vladimir{} Vovk.
\newblock {\em Probability and Finance: It's Only a Game!}
\newblock Wiley, New York, 2001.

\bibitem{shiryaev:1996}
Albert~N\DOT{} Shiryaev.
\newblock {\em Probability}.
\newblock Springer, New York, second edition, 1996.
\newblock Third Russian edition published in 2004.

\bibitem{sorenson:1970}
H\DOT{}~W\DOT{} Sorenson.
\newblock Least-squares estimation: from {G}auss to {K}alman.
\newblock {\em IEEE Spectrum}, 7:63--68, 1970.

\bibitem{steinwart:2001}
Ingo Steinwart.
\newblock On the influence of the kernel on the consistency of support vector
  machines.
\newblock {\em Journal of Machine Learning Research}, 2:67--93, 2001.

\bibitem{GTP14arXiv}
\Vladimir{} Vovk.
\newblock Competitive on-line learning with a convex loss function.
\newblock Technical Report \texttt{arXiv:cs.LG/0506041} (version 3),
  \texttt{arXiv.org} e-Print archive, September 2005.

\bibitem{GTP11arXiv}
\Vladimir{} Vovk.
\newblock On-line regression competitive with reproducing kernel {H}ilbert
  spaces.
\newblock Technical Report
  \texttt{arXiv:\nolinebreak[0]cs.LG/\nolinebreak[0]0511058} (version 2),
  \texttt{arXiv.org} e-Print archive, January 2006.

\end{thebibliography}
\end{document}